\documentclass{article}

\usepackage{PRIMEarxiv}

\usepackage[utf8]{inputenc} 
\usepackage[T1]{fontenc}    
\usepackage{hyperref}       
\usepackage{url}            
\usepackage{booktabs}       
\usepackage{amsfonts}
\usepackage{amsmath}
\usepackage{nicefrac}       
\usepackage{microtype}      
\usepackage{lipsum}
\usepackage{fancyhdr}       
\usepackage{graphicx}       
\graphicspath{{media/}}     
\usepackage{qcircuit}
\usepackage{physics}
\usepackage{xcolor}
\usepackage{longtable}
\usepackage{multirow}
\usepackage[nottoc]{tocbibind}
\usepackage{subcaption}

\usepackage{algorithm}
\usepackage{algorithmic}

\title{Synthetic Data in Cryptocurrencies using Generative Models 
\thanks{\textit{\underline{Citation}}: 
\textbf{Sousa A. S. S. et al. Synthetic Data in Finance using Generative Models. Pages.... DOI:000000/11111.}} 
}

\author{
André Saimon S. Sousa\\
Universidade SENAI CIMATEC\\
Salvador, Brazil\\
\texttt{andre.sousa@fbter.org.br} \\
\And
Otto Pires\\
Universidade SENAI CIMATEC\\
Salvador, Brazil\\
\texttt{otto.pires@fieb.org.br} \\
\And
Frank Acasiete\\
Universidade SENAI CIMATEC\\
Salvador, Brazil\\
\texttt{frank.quispe@fbter.org.br} \\
\And
Oscar M. Granados\\
CoFi, Universidad Jorge Tadeo Lozano\\
Bogotá, Colombia\\
\texttt{oscarm.granadose@utadeo.edu.co} \\
\And
Valéria Loureiro da Silva \\
Universidade SENAI CIMATEC\\
Salvador, Brazil\\
\texttt{valeria.dasilva@fieb.org.br}
\And
Hugo Saba \\
Universidade do Estado da Bahia - UNEB \\
Salvador, Brazil\\
\texttt{hugosaba@uneb.br}
}

\begin{document}
\maketitle

\begin{abstract}
Data plays a fundamental role in consolidating markets, services, and products in the digital financial ecosystem. However, the use of real data, especially in the financial context, can lead to privacy risks and access restrictions, affecting institutions, research, and modeling processes. Although not all financial datasets present such limitations, this work proposes the use of deep learning techniques for generating synthetic data applied to cryptocurrency price time series. The approach is based on Conditional Generative Adversarial Networks (CGANs), combining an LSTM-type recurrent generator and an MLP discriminator to produce statistically consistent synthetic data. The experiments consider different crypto-assets and demonstrate that the model is capable of reproducing relevant temporal patterns, preserving market trends and dynamics. The generation of synthetic series through GANs is an efficient alternative for simulating financial data, showing potential for applications such as market behavior analysis and anomaly detection, with lower computational cost compared to more complex generative approaches.
\end{abstract}


\section{Introduction}

Financial data has grown in several ways. In some cases, these data are exclusive to financial institutions, and they prefer not to share them with other institutions for regulatory or strategic reasons. Open finance and open data are realities for specific data, but transactions remain secret between institutions. Transaction records have been a highlight in identifying anomalies such as volatility risk, operational risk, fraud, financial crime, and money laundering. However, the models are habitually endogenous with data from each institution.  This obstacle affects the deployment of methods based on artificial intelligence. Thus,  synthetic data emerges as a solution to address this challenge on different financial datasets, which can further refine the training of deep reinforcement learning models used in money laundering control, computational finance, or algorithmic and high-frequency trading.

Our problem setting involves applying generative AI to enhance anomaly detection capabilities with synthetic data. Generative AI, particularly GANs, is employed to create synthetic data that closely resembles real-world data. These synthetic data serve as a valuable resource for training and testing anomaly detection models. By augmenting existing datasets with synthetic data, we can address issues such as data scarcity, privacy concerns, and imbalanced class distributions. This approach enables the development of more robust and accurate anomaly detection systems such as volatility risk, hedging exposition, financial crime, money laundering, etc. Early research led to the development of a synthetic information generator for use in both public and private sector research. The proposal consisted of multiple imputations, i.e., conjugating and simulating with probabilistic sampling techniques called Bootstrap and carrying out simulations on sets of samples with or without replacement to estimate the value~\cite{Rubin}. The estimated value was listed as the synthetic value with no record. Furthermore, another proposal was to find highly correlated auxiliary variables to replace the target variable. Subsequently, other techniques were developed to create synthetic values through multiple imputations using point estimates, supervised and unsupervised learning models, and deep learning~\cite{Rubin}.

Consequently, GANs are a class of machine learning frameworks that consist of two neural networks competing in a zero-sum game: a generator and a discriminator. The generator's goal is to create new data instances that are indistinguishable from real data, while the discriminator's goal is to accurately classify these instances as real or fake~\cite{Radford}. During training, the generator produces synthetic data, and the discriminator evaluates its authenticity. The discriminator provides feedback to the generator, which then adjusts its parameters to improve the realism of its generated data. This adversarial process continues iteratively, with both networks learning from each other~\cite{Goodfellowetal2014}.
Over time, the generator becomes increasingly skilled at creating highly realistic synthetic data, while the discriminator's ability to distinguish between real and fake data improves. This dynamic competition drives the development of sophisticated generative models capable of producing diverse and convincing outputs.

GANs have found a wide range of applications across various domains, including finance. Like Synthetic Data Generation, privacy-preserving Data: GANs can generate synthetic financial data that mimics real-world data but preserves privacy by removing sensitive information. Data augmentation, by generating synthetic data, GANs can augment existing datasets, improving the performance of machine learning models. Stress testing, GANs can simulate various economic scenarios to stress-test financial models and risk management strategies.

Compared to purely mathematical linear models, GANs are able to present better results in representing complex distributions related to cryptocurrency price volatility, preserving statistical characteristics and allowing the generation of realistic data ~\cite{jrfm15010026}. The generation of synthetic data using GANs also stands out for presenting advantages in enriching datasets in cryptocurrency contexts, increasing the accuracy of predictive models, reducing overfitting, improving robustness, and increasing accuracy ~\cite{jrfm15010026, Almamoori2023, 10246252}.

GANs also have applications in various areas. For example, Zhu~\cite{Zhu} developed an algorithm for realistic photographic manipulation of shape and color. Killoran~\cite{Killoran} proposed generative neural network methods to create adjustable DNA sequences. Additionally, Kadurin~\cite{Kadurin} developed a model for molecular feature extraction problems.
GANs\cite{salimans2016} stand out as a powerful tool for data generation. They have gained significant traction in the field of generative learning and find application in an extensive variety of domains\cite{Bojchevski2018, ngo2023survey, Dash2024}. In the finance domain, applications of GANs include financial data generation\cite{Takahashi2019121261, Naritomi2020641,Zhang2025103}, stock market prediction\cite{zhou2018stock, zhang2019stock, li2020generating, Vuletić01022024}, credit scoring\cite{Kang2022117650}, fraud detection\cite{leangarun2018stock, Fiore2019448, Aftabi2023120144}, and money laundering \cite{Granadosetal2024}. Several other efforts have focused on leveraging deep learning methodologies, particularly in medical data \cite{IoT,doi:10.1161/CIRCOUTCOMES.118.005122, Sandfort2019, Jordon2019PATEGANGS}. 

This work aims to gain a better understanding of financial anomalies by using synthetic data methods with GAN networks. Indeed, we apply some tools from diffusion models that are adapted to this setting to create data and detect the most relevant financial anomalies that develop interactions that are frequently associated with several financial issues. For this purpose, we deploy a model to create synthetic data in cryptocurrencies as a basis for the analysis of other financial datasets and introduce a proposal for the anomaly detection of financial transactions and operations in high volatility contexts. 
We will also employ other deep learning methods as an aid in this study, other deep learning methods to consolidate our model, which, in the case of financial data, has the advantage of having an interpretation and is well-defined in settings where volatility increases suddenly. It is desirable for a more refined analysis of anomalies, but it will not be attempted here since our purpose is to consider the case when only the minimum quantity of information is available.

This work has two main contributions. First, provide clarity on the evolution of the dynamics and structure of the three selected cryptocurrencies across different volatility periods, and assess whether properties inherent to these cryptocurrencies and to the trading periods could have imposed limitations on the creation of synthetic data. Second, to extend the understanding of these periods from a machine learning perspective by (a) creating a deep learning model with combine two methods to directly create dataset by minute that built transactions in all cryptocurrencies in each market, and (b) by providing an extensive experiments of the temporal and volatility changes in the transactions of each of these markets during the overall period that consolidate a synthetic data with all properties of real markets.

This paper is divided as follows. In Section~\ref{sec:background}, we present some basic neural networks and GAN network terminology employed in the text. The Sect. 3 is devoted to a brief description of the datasets used in this study, and a basic data analysis also describes our method. In Section~\ref{sec:results}, we describe our experiments and analyze the results. Finally, we close with a conclusion and future works in Section~\ref{sec:conclusion}.

\section{Preliminaries}\label{sec:background}

Given observed samples x from a distribution of interest, the goal of a generative model is to learn to model its true data distribution $p(x)$. Once learned, we can generate new samples from our approximate model at will. Furthermore, we can use the learned model to evaluate the likelihood of observed or sampled data as well. Below, some concepts are developed to understand our proposed models.

\subsection{Neural Networks}

Artificial Neural Networks (ANNs) are an area of Artificial Intelligence where the structure of their model seeks to resemble the behavior of biological neurons and, by using many layers of complex algebraic circuits, is called deep learning \cite{ferreira2021deep}. The perceptron, which forms the basis of this system, can be described as a node, also called a unit, that performs the calculation of the weighted sum of the inputs from the previous nodes and then applies a linear function in order to generate an output \cite{russell2021ia}. The output of unit $j$, represented by $\textbf{a}_j$, is given by the activation function $\textbf{g}_j$ applied to the weighted sum of the inputs $i$, as expressed in the Eq.~\eqref{eq:RNA}.

\begin{equation} 
\label{eq:RNA} 
\begin{aligned}
\mathbf{a}_j = \mathbf{g}_j \left(\sum_i \mathbf{w}_{i,j} \mathbf{a}_i \right) \equiv \mathbf{g}_j(\mathbf{in}_j)
\end{aligned}
\end{equation}

where $\textbf{w}_{i,j}$ refers to the weight of the link from $i$ to $j$ and $\textbf{in}_j$ refers to the weighted sum \cite{russell2021ia}. It is important to highlight that ANN architectures have specific characteristics for certain classes of problems, such as the case where convolutional networks generalize well to spatial grids (images), while recurrent networks are better suited to sequential data flows \cite{russell2021ia}.

For the use of sequential data, Recurrent Neural Networks (RNNs) are ideal because they establish cycles in the computation graph, allowing for memory states and the storage of temporal dependencies \cite{russell2021ia}. In an RNN, at each time interval, the input and output are observed, and the memory state is established through the recurrence of information processing by the hidden layer \cite{russell2021ia}, as demonstrated in the Eq.~\eqref{eq:RNN}.

\begin{equation} 
\label{eq:RNN} 
\begin{aligned}
\mathbf{z}_t = \mathbf{f}_\mathbf{w}(\mathbf{z}_{t-1},\mathbf{x}_t) = \mathbf{g}_z(\mathbf{W}_{z,z} \mathbf{z}_{t-1} + \mathbf{W}_{x,z} \mathbf{x}_t \equiv \mathbf{g}_z (\mathbf{in}_{z,t}) \\
\hat{\mathbf{y}}_t = \mathbf{g}_y ( \mathbf{W}_{z,y} \mathbf{z}_t) \equiv \mathbf{g}_y( \mathbf{in}_{y,t})
\end{aligned}
\end{equation}

The Long Short-Term Memory (LSTM) network follows the same principles as an RNN, adding memory cells to the network capable of preserving information for many time intervals \cite{russell2021ia}. For the equation below (Eq.~\eqref{eq:LSTM}), $\textbf{c}$ refers to the switching component that controls the flow of information; the forget gate $\textbf{f}$ defines what will be forgotten and what will be remembered; the input gate $i$ defines which information will be added; the output gate $\mathbf{o}$ organizes the short-term memory $z$; the subscripted weight matrices $\mathbf{W}$ refer to the origin and destination; and the symbol $\odot$ represents element-by-element multiplication.

\begin{equation} 
\label{eq:LSTM} 
\begin{aligned}
\mathbf{f}_t = \sigma (\mathbf{W}_{x,f} \mathbf{x}_t + \mathbf{W}_{z,f} \mathbf{z}_{t-1}) \\
\mathbf{i}_t = \sigma (\mathbf{W}_{x,i} \mathbf{x}_t + \mathbf{W}_{z,i} \mathbf{z}_{t-1}) \\
\mathbf{o}_t = \sigma (\mathbf{W}_{x,o} \mathbf{x}_t + \mathbf{W}_{z,o} \mathbf{z}_{t-1}) \\
\mathbf{c}_t = \mathbf{c}_{t-1} \odot \mathbf{f}_t + \mathbf{i}_t \odot \text{tanh}(\mathbf{W}_{x,c}\mathbf{x}_t + \mathbf{W}_{z,c}\mathbf{z}_{t-1}) \\
\mathbf{z}_t = \text{tanh}(\mathbf{c}_t) \odot \mathbf{o}_t
\end{aligned}
\end{equation}

\begin{figure}[H]
\centering
\begin{subfigure}{0.95\textwidth}
\includegraphics[width=\textwidth]{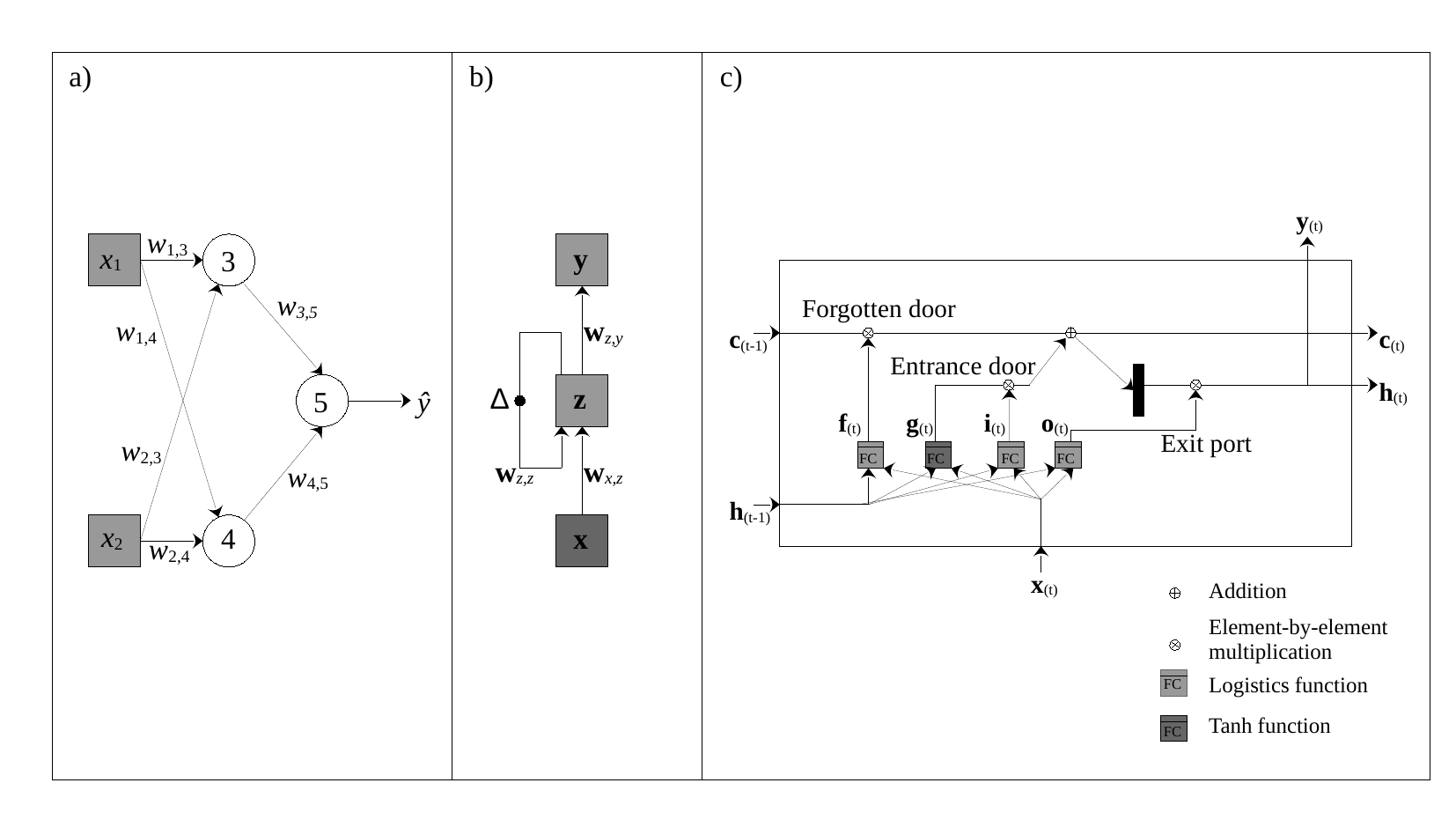}
\caption{Neural Networks: a) ANN; b) RNN; c) LSTM.}
\label{fig:RNA}
\end{subfigure}
\end{figure}

Figure \ref{fig:RNA} shows an example of the basic structure of an ANN, an RNN, and an LSTM, respectively, from left to right \cite{russell2021ia} \cite{geron2021maos}.

Fig.~\ref{fig:RNA} 

\subsection{Generative Adversarial Networks (GAN)}
The GAN networks involve two competing models of ANNs. The initial model, denoted as the generator $(g)$, is trained to generate synthetic data, while the second model, the discriminator $(d)$, is trained to distinguish between fake and real data. In this framework, a zero-sum cooperative game is played, in which both players attempt to optimize their strategies to outperform their adversary. This specific context focuses on the two neural networks, ensuring a high level of efficiency in learning by achieving convergence in the cost loss function between them. The generator refines its learning based on the feedback received from the discriminator's classifications~\cite{LangrBok2019}. The discriminator's objective is to determine whether the data is real (from the training dataset) or false (created by the white noise generator). 

We can represent the scenario where the game competition function $ \left(\min; \max \right) $ is expressed between the two networks with their goodness parameters of each one of them $ v\left( \theta^{(g)};\theta^{(d)} \right) $ Eq.~\eqref{eq:minmax}. That is, during learning, each player tries to maximize his reward, so that the convergence is given by:

\begin{equation} 
\label{eq:minmax} 
 g^{\ast} = \arg \min_{g}  \max_{d} v\left(g,d\right).
\end{equation}

The function is defined by {\large{$f\left(x,v\left(g,d\right)\right)$}} the Eq.~\eqref{eq:Gans}:  

\begin{equation} \label{eq:Gans}
\begin{split}
 v\left(\theta^{(g)},\theta^{(d)}\right) = \mathbb{E}_{x \sim p_{\text{data}}}(x) \left[ \log d(x)\right] + \mathbb{E}_{z \sim p_z}(z) \left[ \log \left(1-d\left( g\left(z \right) \right) \right)\right],
\end{split}
\end{equation}

where: $\left(x\right)$ represents the real training data, $p_{\text{data}} (x)$ of their distribution, $ \left(z \right) $ is noise with which the generator is fed to synthesize the data, and its distribution $ p_{z}(z) $ which is determined as the distribution of the generated data $p_{g}(z)$. The first neural network, called the \textit{Generator} $ (g) $, consists of generating samples given at $ x = g \left(z; \theta ^ {(g)} \right) $. Its adversary, the second neural network named \textit{Discriminator} $ (d) $, tries to distinguish between samples drawn from the training data and samples drawn from the generator. The discriminator outputs a probability value given by $ d \left (x; \theta^{(d)} \right) $, which calculates or estimates a probability of $ x $ that a sample comes from the training data instead of a fake sample taken from the $(g) $ \cite{Goodfellowetal2016} model.

Then, learning occurs through a zero-sum game, in which a function given by $ v (\theta ^{(g)}, \theta ^{(d)}) $ determines the reward of the discriminator~\cite{Goodfellowetal2016}. Also, the generator receives $ -v (\theta^{(g)}, \theta^{(d)}) $ as its reward, so the discriminator learns to classify the samples as real or false correctly. Simultaneously, the generator tries to trick the discriminator into believing that its samples are real. In addition, the convergence of the generator samples is indistinguishable from the real data, which means that the discriminator must have a high level of learning with a minimum error rate and maximize the probabilities for the discriminator to make estimates of high performance given at the learning task as represented by Eq.~\eqref{eq:Gans}. Additionally, the generator ($g$) implicitly defines a probability distribution $ p_{g} $ as the distribution of the samples $ g (z) $ obtained when $z \sim p_{z} $~\cite{Goodfellowetal2014}. Therefore, it is sought that $(g)$ converges to a good unbiased estimator for $ p_{\text{data}} $. 

Generative models can be classified into GANs, Variational Autoencoders (VAEs), and AutoRegressive Networks \cite{8667290}. GAN is a Machine Learning model presented in 2014 that advances the proposal of generative models by using backpropagation to optimize the network weights, in contrast to the use of Markov chains and approximate inference in previous models \cite{goodfellow2014}. GANs aim to indicate, through a discriminating network $D$ and probability estimation, whether the data distribution is real or created through the generating network $G$, functioning as a minimax game in its cost function.

VAE consists of a probabilistic graphical model that aims to model the probabilistic distribution of data, but with a certain bias, generating samples with lower quality than GANs. Regarding AutoRegressive Networks, PixelRNN is an example that performs pixel prediction and image generation by processing the pixels one by one, while GANs process the sample all at once, making it more efficient \cite{8667290}. 

The generator is trained so that the probability of the discriminator making an error is maximized, while the discriminator aims to learn whether the data sample comes from the model distribution or from the data distribution \cite{goodfellow2014}. In its initial proposal, the generative model creates samples by passing random noise through a multilayer perceptron, and the discriminative model is also a multilayer perceptron \cite{goodfellow2014}.

The game should be implemented using an iterative numerical approach, optimizing $D$ by alternating between $k$ steps and $G$ at each step, keeping $D$ close to its optimal solution by slowly changing $G$ \cite{goodfellow2014}. The training objective for $D$ can be interpreted as maximizing the log-likelihood to estimate the conditional probability $P(Y = y|x)$, where $Y$ indicates whether $x$ comes from $p_{data}$ (with $y = 1$) or from $p_g$ (with $y = 0$) \cite{goodfellow2014}.

The generator takes as input a random noise vector $z$, which is typically drawn from a uniform or normal distribution. To obtain a multidimensional vector, which is a fake sample $G(z)$, the noise is mapped to a new data space through the generator. Then, the discriminator receives from the data set the real sample and the fake sample created by the generator, and delivers an output representing the probability characterizing the sample as real, instead of fake, acting as a binary classifier. The discriminator reaches the optimal state when it learns the distribution of the real data and makes it indistinguishable to classify the data as coming from the real or fake sample \cite{8667290}.

\section{Materials and Methods} \label{sec:materials}

Several aspects are critical when developing an Artificial Neural Network (ANN) model. It has four fundamental components: (a) the dataset; (b) the data transformation model; (c) an objective function to evaluate model quality; and (d) a tuning algorithm for optimization \cite{zhang2023dive}. The generation of synthetic data offers a methodological alternative to the manual labeling and protection of sensitive information, such as financial records. Consequently, synthetic data facilitates the generation of new samples while preserving the underlying relationships and attributes of the original dataset. This data is typically classified into three categories. First, completely synthetic data, where a generator constructs values for each variable based on probabilistic parameters. Second, partially synthetic data, which uses statistical imputation and dimension reduction to transform information. Third, hybrid synthetic data, which integrates elements from both real and generated sources.

\begin{figure}[H]
\centering
\begin{subfigure}{0.95\textwidth}
\includegraphics[width=\textwidth]{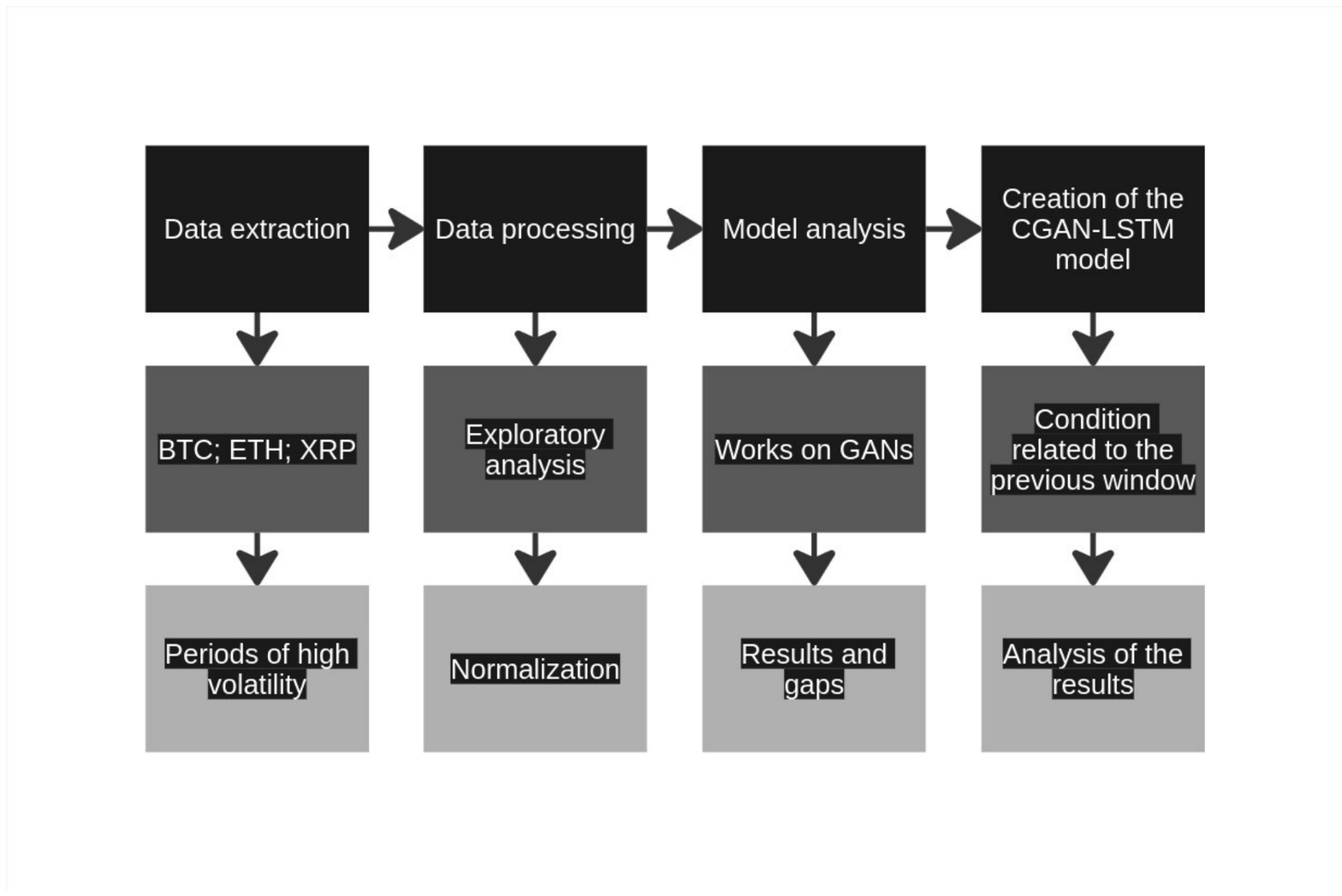}
\caption{Methodological Flowchart}
\label{fig:method}
\end{subfigure}
\end{figure}

Within the framework of this study (Fig \ref{fig:method}), the process begins by extracting cryptocurrency data (BTC, ETH, and XRP) over a specific temporal window. Data preprocessing involves cleaning and normalizing these values based on an analysis of the time-series characteristics. To develop the GAN architecture, we conducted a review of existing models to identify best practices for construction and performance evaluation. Finally, we propose a model specifically tailored to the structure of the available data.

Fig.~\ref{fig:method}.

\subsection{Data}

The general period covered in this work spans January 2022 to October 2025. We study cryptocurrencies using minute-by-minute data across three time points. First, from March 2022 to April 2022. Second, from April 2022 to May 2022. Third, from September 2025. This work focuses on these three periods starting from the 21$^{st}$ of January 2022 (when the attack of Russia on Ukraine was confirmed) and the one subsequent month. This period was selected as it portrays the beginning of the war in Europe, as well as when the implementation of monetary policies of several countries started to reorganize the level of interest rates, announced in the last days of March and early April of 2022. The final period refers to the global trade changes with Trump's strategy. This work refers to these three periods as the ``Volatility scenario'' or ``VS period'' for short throughout this document.  The dataset comes from the LSEG Platform and contains approximately 0.5 million records covering three currencies: Bitcoin (BTC), Ethereum (ETH), and XRP. We impose a strict survivorship filter and retain only records with all information and present over the entire sample. This harmonizes the cross-section through time and ensures that correlation-based data are well defined. 

The records have five variables: period, open, high, low, and close. For data preprocessing, we considered several criteria regarding the quality. The data eliminated were records with some anomaly. In addition, the month, day, and time data were extracted from the date field, thus creating three additional variables to become factors in the analysis of the transactions. For the final dataset, we have 384465 records. 

\subsection{Data Analysis}

\begin{figure}
    \centering
    \includegraphics[width=0.99\linewidth]{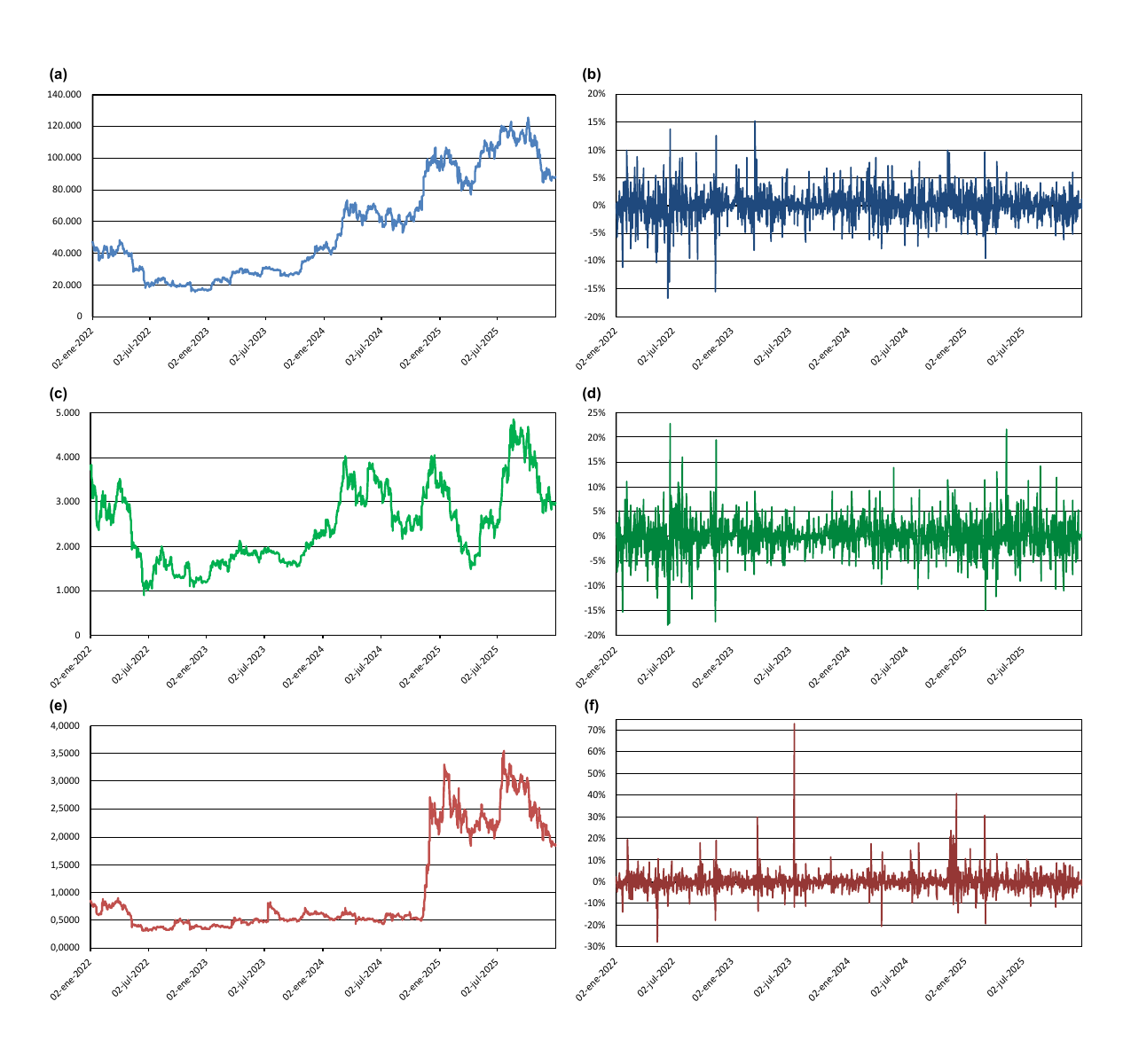}
    \caption{Cryptocurrencies evolution (2022-2025)}
    \label{fig:3}
\end{figure}

\begin{figure}
    \centering
    \includegraphics[width=0.99\linewidth]{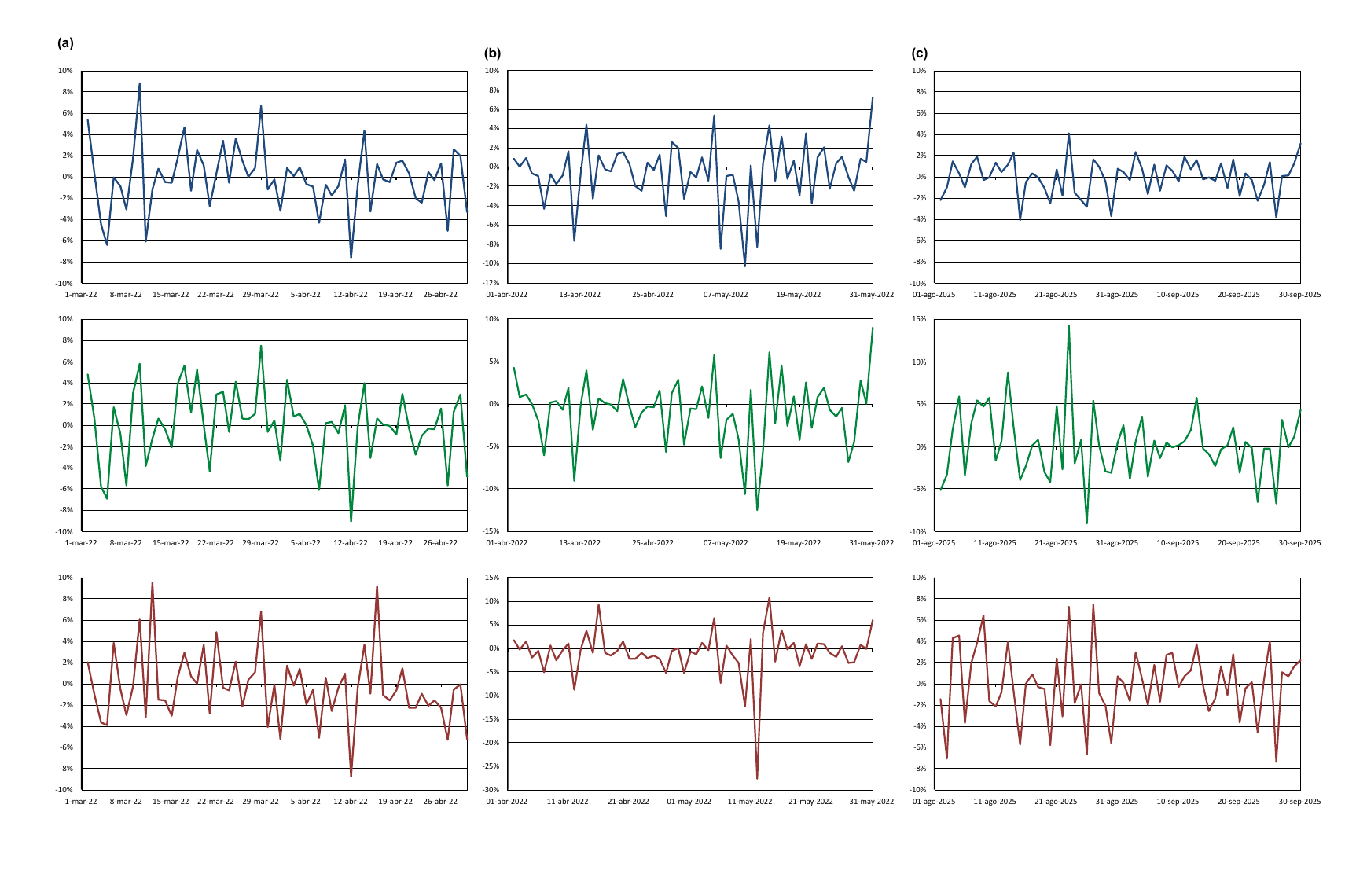}
    \caption{Volatility by periods}
    \label{fig:4}
\end{figure}

The dataset, segmented into three distinct temporal windows (Fig. \ref{fig:4}), illustrates the nature of cryptocurrency price fluctuations during periods of significant volatility, driven by macroeconomic or geopolitical events. By analyzing the variance of daily percentage changes, we can identify the market's sensitivity to exogenous shocks and the subsequent risk inherent in financial assets, including digital assets.
During the first period, March–April 2022, the charts in column (a) (Fig. \ref{fig:4}a) reflect a market grappling with the immediate fallout of the Russian invasion of Ukraine. This period is characterized by high-frequency oscillations, with daily changes frequently swinging between +8\% and -8\%. This represents a risk-off sentiment where investors oscillated between viewing Bitcoin or other cryptocurrencies as an asset hedge, but at the same time, a speculative high-beta asset. The intraday volatility seen throughout March 2022 indicates several perspectives on cryptocurrency trends, as geopolitical sanctions and energy supply concerns induced rapid shifts in financial assets' liquidity.
Column (b) represents the most difficult phase of the observed timeline of the three datasets, coinciding with the change of the US monetary policy and geopolitical issues, which led to the fall of several assets and the collapse of major algorithmic stablecoin ecosystems. The volatility here is not a result of an oscillatory process but directional and precipitous. We observe extreme negative outliers, with some assets plunging beyond -15\% and even touching the -30\% threshold by mid-May 2022. This period highlights the systemic risk and structural fragility, exposing underlying structural vulnerabilities in these kinds of financial assets. The charts illustrate a traditional feedback loop where margin calls and forced liquidations amplified the downward pressure, leading to a general shift in market confidence.
The data for 2025 in column (c) suggests a transition toward a more mature, albeit still volatile, market structure. While geopolitical tensions remained a persistent factor, the amplitude of the daily variations narrowed compared to each 2022 period. Most fluctuations were contained within a ±4\% range, though occasional spikes to +10\% or -10\% persist. The 2025 charts show a market that had priced in a portion of geopolitical and macroeconomic risk into its pricing model, exhibiting a more "dampened" response to external shocks compared to the behavior of 2022. However, volatility consistently re-emerges during periods of geopolitical and macroeconomic uncertainty.

\subsection{Method} \label{method}

\begin{figure}[H]
\centering
\begin{subfigure}{0.95\textwidth}
\includegraphics[width=\textwidth]{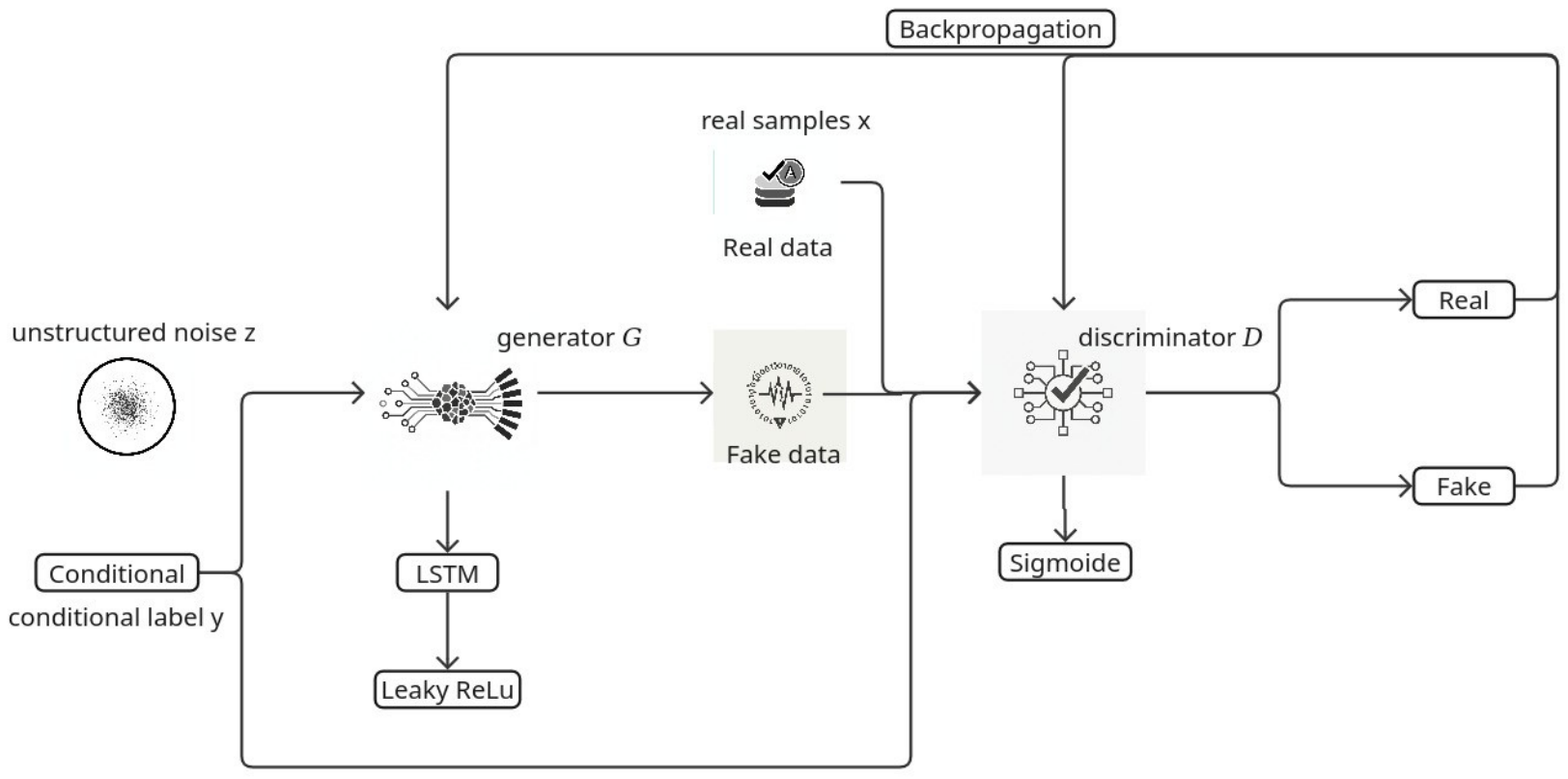}
\caption{Architecture}
\label{fig:architecture}
\end{subfigure}
\end{figure}

The defined architecture is a Conditional Generative Adversarial Network (CGAN) with a hybrid LSTM-MLP architecture, illustrated by Fig \ref{fig:architecture}. Processing begins with data loading, conversion from the original format to .csv, chronological ordering, and normalization of values using the StandardScaler method. 

Data normalization was performed using the StandardScaler method, widely used in machine learning tasks for standardizing numerical variables. This method transforms the data so that they have a mean of zero and a unit variance, as shown in the following expression (Eq.~\eqref{eq:StandardScaler}).

\begin{equation} \label{eq:StandardScaler}
\begin{split}
x^{\prime} = \frac{x - \mu}{\sigma},
\end{split}
\end{equation}

where \( x \) represents the original value of the observation, \( \mu \) corresponds to the sample mean of the series, and \( \sigma \) to the standard deviation. This transformation ensures that the values are centered around the origin and scaled uniformly, preserving the shape of the original data distribution.

In the context of this research, the StandardScaler was applied exclusively to the cryptocurrency closing variable, allowing both the generator and the discriminator to operate on normalized data during training. After the generation process, the inverse transformation was applied to recover the values at the original scale, thereby enabling economic interpretation of the synthetic results.

The data are structured as pairs, linking values with a conditional label referring to the previous value, $(Price_{t-1}, Price_t)$. When training the network, the Generator (LSTM) receives the previous price and noise to generate the current price ($Price_t$), and the Discriminator (MLP) classifies the pairs as real or synthetic. The figure below shows the defined architectural format.

Fig.~\ref{fig:architecture}.

We are going to use the following algorithm based on the one proposed by em~\cite{Saravana}
\begin{algorithm}[H]
\caption{Conditional GAN for Time Series Prediction}
\label{alg:cgan}
\begin{algorithmic}[1]
\STATE \textbf{Input:} Time series of closing prices, noise dimension $l$, condition dimension $d$, batch size $k$, number of epochs $E$, learning rate $\eta$, Adam optimizer parameters $(\beta_1, \beta_2)$.
\STATE \textbf{Output:} Trained parameters of Generator $G$ and Discriminator $D$.

\STATE \textbf{Data Loading and Preprocessing:}
\STATE Load closing price time series.
\STATE Normalize the data using \texttt{StandardScaler}.
\STATE Create training pairs $(y_i, x_i)$, where the condition $y_i$ is the normalized price at time $t-1$ and the real data $x_i$ is the normalized price at time $t$.

\STATE \textbf{Initialization:}
\STATE Set hyperparameters: $l=8$, $d=60$, $k=64$, $E=50$, $\eta=0.0002$, $\beta_1=0.5$, $\beta_2=0.999$.
\STATE Initialize neural networks:
\STATE $G(y, z)$: LSTM receiving condition $y$ and noise vector $z$ (internally generated) to output $\hat{x}$ (predicted price at $t$).
\STATE $D(y, x)$: MLP receiving condition $y$ and value $x$ (real or generated) to return classification logits (real vs. fake).
\STATE Initialize Adam optimizers for $G$ and $D$ with the defined hyperparameters.
\STATE Initialize loss function: Binary Cross-Entropy with Logits (\texttt{BCEWithLogitsLoss}).

\STATE \textbf{Training Loop:}
\FOR{each epoch $e \in \{1, \ldots, E\}$}
  \FOR{each batch $(y, x_{\text{real}})$ from the dataset}
    \STATE \textbf{Train Discriminator:}
    \STATE Sample real data $x_{\text{real}}$ with corresponding conditions $y$.
    \STATE Compute real validity: $logits_{\text{real}} = D(y, x_{\text{real}})$.
    \STATE Compute real loss: $L_D^{real} = \text{BCEWithLogitsLoss}(logits_{\text{real}}, \mathbf{1})$.
    \STATE Generate fake data: $\hat{x} = G(y)$ (noise $z$ is generated inside $G$).
    \STATE Compute fake validity (no gradients for $G$): $logits_{\text{fake}} = D(y, \hat{x}.detach())$.
    \STATE Compute fake loss: $L_D^{\text{fake}} = \text{BCEWithLogitsLoss}(logits_{\text{fake}}, \mathbf{0})$.
    \STATE Compute total discriminator loss: $L_D = \tfrac{1}{2}(L_D^{\text{real}} + L_D^{\text{fake}})$.
    \STATE Update discriminator weights: $\theta_D \leftarrow \text{Adam}(\theta_D, \nabla_{\theta_D} L_D)$.

    \STATE \textbf{Train Generator:}
    \STATE Generate synthetic sequences: $\hat{x} = G(y)$.
    \STATE Evaluate with Discriminator: $logits_{\text{gen}} = D(y, \hat{x})$.
    \STATE Compute generator loss (to fool the Discriminator): $L_G = \text{BCEWithLogitsLoss}(logits_{\text{gen}}, \mathbf{1})$.
    \STATE Update generator weights: $\theta_G \leftarrow \text{Adam}(\theta_G, \nabla_{\theta_G} L_G)$.
  \ENDFOR
\ENDFOR
\end{algorithmic}
\end{algorithm}

The Binary Cross-Entropy with Logits (BCEWithLogitsLoss) cost function is widely used in binary classification problems and constitutes a numerically stable formulation of binary cross-entropy combined with the sigmoid function. Instead of explicitly applying the sigmoid function to the model output and then calculating the cross-entropy, this function integrates both operations into a single analytical expression, reducing numerical errors associated with extreme exponential operations.

Formally, considering a scalar output of the discriminator \( x \in \mathbb{R} \) (logit) and a binary label \( y \in \{0,1\} \), the cost function is defined as (Eq.~\eqref{eq:BCEWithLogitsLoss}):

\begin{equation} \label{eq:BCEWithLogitsLoss}
\begin{split}
\mathcal{L}(x, y) = - \left[ y \cdot \log\big(\sigma(x)\big) + (1 - y) \cdot \log\big(1 - \sigma(x)\big) \right],
\end{split}
\end{equation}

where \( \sigma(\cdot) \) represents the sigmoid function, defined by the Eq.~\eqref{eq:sigma}:

\begin{equation} \label{eq:sigma}
\begin{split}
\sigma(x) = \frac{1}{1 + e^{-x}}.
\end{split}
\end{equation}

The computational implementation uses an equivalent reformulation based on the \textit{log-sum-exp} technique, which avoids numerical saturation when \( |x| \) assumes high values \cite{kingma2015adam, paszke2019pytorch}.

The validation of the generated data is done through the use of performance metrics, such as Pearson's correlation, in order to evaluate statistical similarities between the generated and real series, in addition to MAE (Mean Absolute Error) and RMSE (Root Mean Square Error). Complementarily, visual verification through comparisons shown in graphs allows visualization of whether the model correctly captured the behavior of the data and the dynamics of the established series.

\section{Results and Discussion}\label{sec:results}

The figures presented below demonstrate the results of the Conditional GAN-LSTM model's behavior when applied to minute-by-minute closing price data for BTC, ETH, and XRP across the three selected time periods. The total number of records for each period is shown below, and 50 epochs were used for each training session.

First period (from March 2022 to April 2022): BTC - 22818; ETH - 22820; XRP - 22821.

Second period (from April 2022 to May 2022): BTC - 84428; ETH - 85693; XRP - 85502.

Third period (from September 2025): BTC - 20124; ETH - 20130; XRP - 20129.

\begin{figure}[H]
\centering

\begin{subfigure}{0.55\textwidth}
\includegraphics[width=\textwidth]{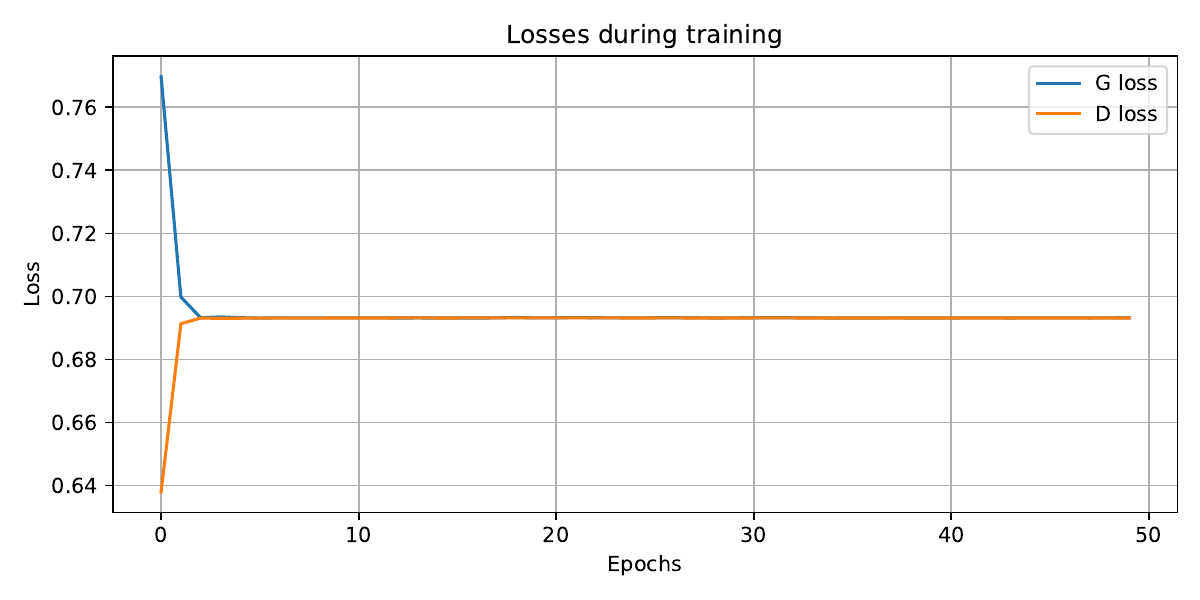}
\caption{BTC - data first period.}
\label{fig:lossbtcData1}
\end{subfigure}

\begin{subfigure}{0.55\textwidth}
\includegraphics[width=\textwidth]{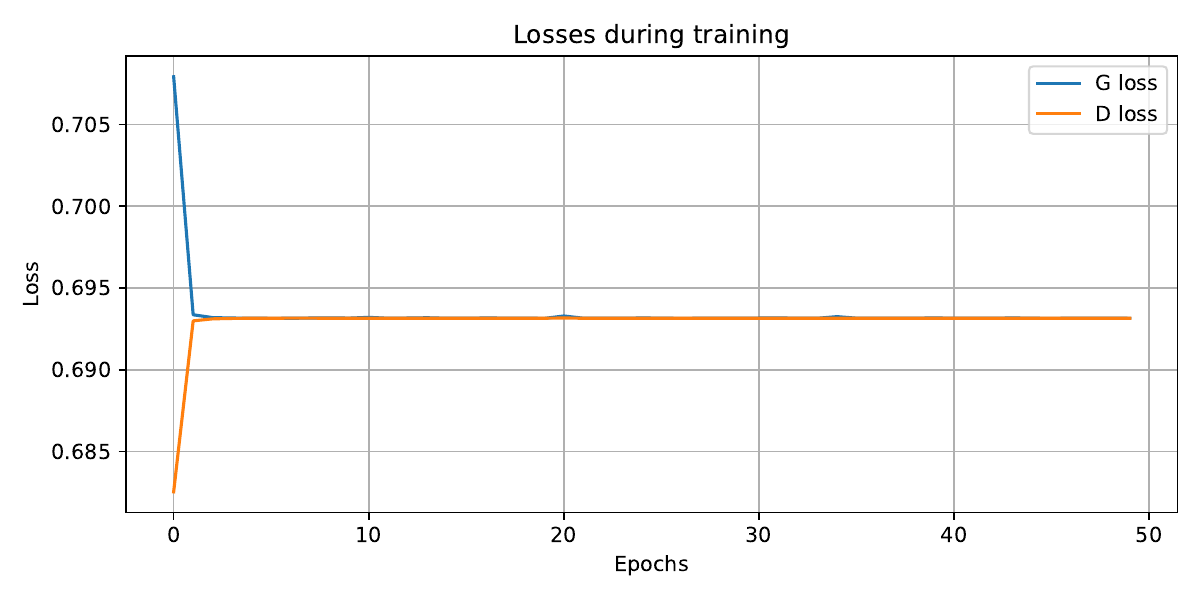}
\caption{BTC - data second period.}
\label{fig:lossbtcData2}
\end{subfigure}

\begin{subfigure}{0.55\textwidth}
\includegraphics[width=\textwidth]{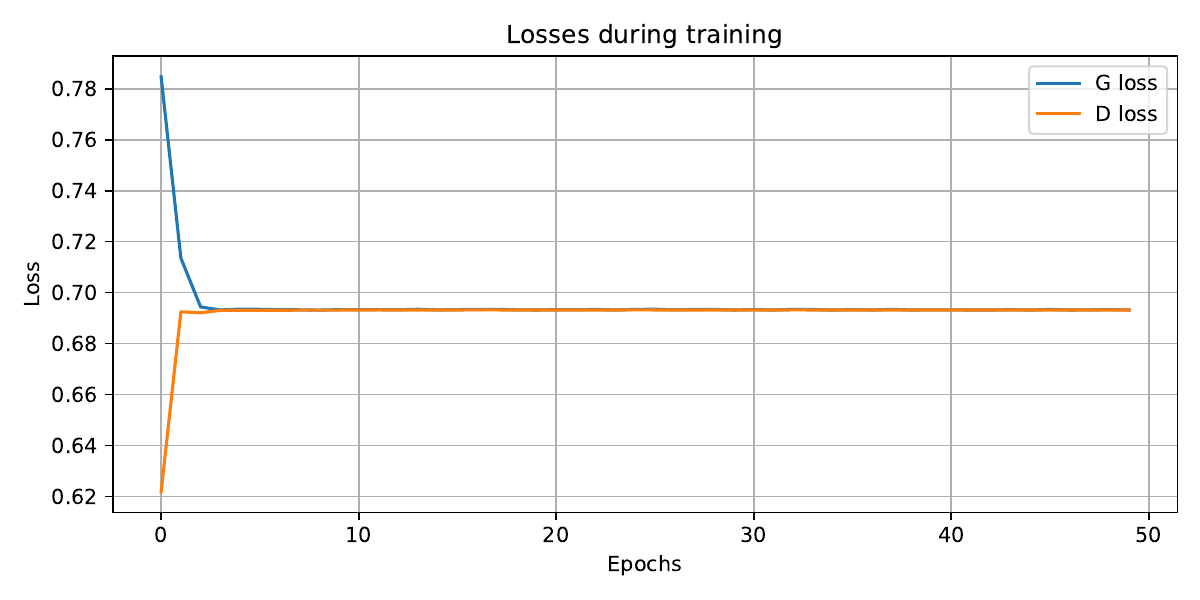}
\caption{BTC - data third period.}
\label{fig:lossbtcData3}
\end{subfigure}

\caption{Losses during training: BTC.}
\label{fig:lossbtcData}
\end{figure}

The Fig.~\ref{fig:lossbtcData} shows a comparison between the loss values of the generating and discriminating networks for the first, second, and third periods of BTC values.

\vspace*{-0.5cm}

\begin{figure}[H]
\centering

\begin{subfigure}{0.55\textwidth}
\includegraphics[width=\textwidth]{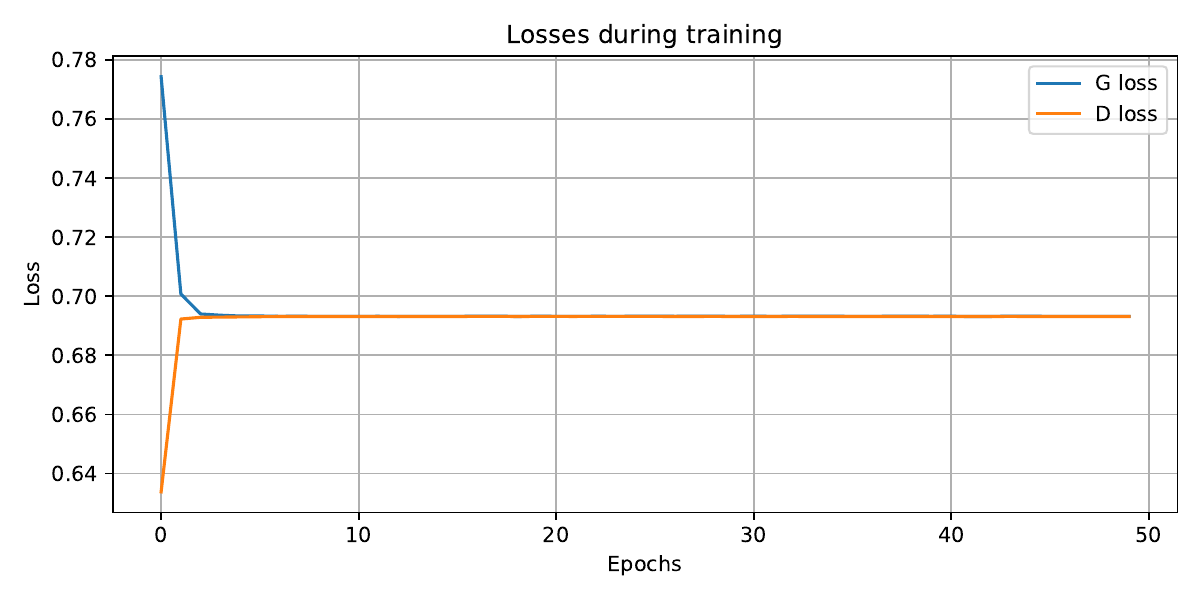}
\caption{ETH - data first period.}
\label{fig:lossethData1}
\end{subfigure}

\begin{subfigure}{0.55\textwidth}
\includegraphics[width=\textwidth]{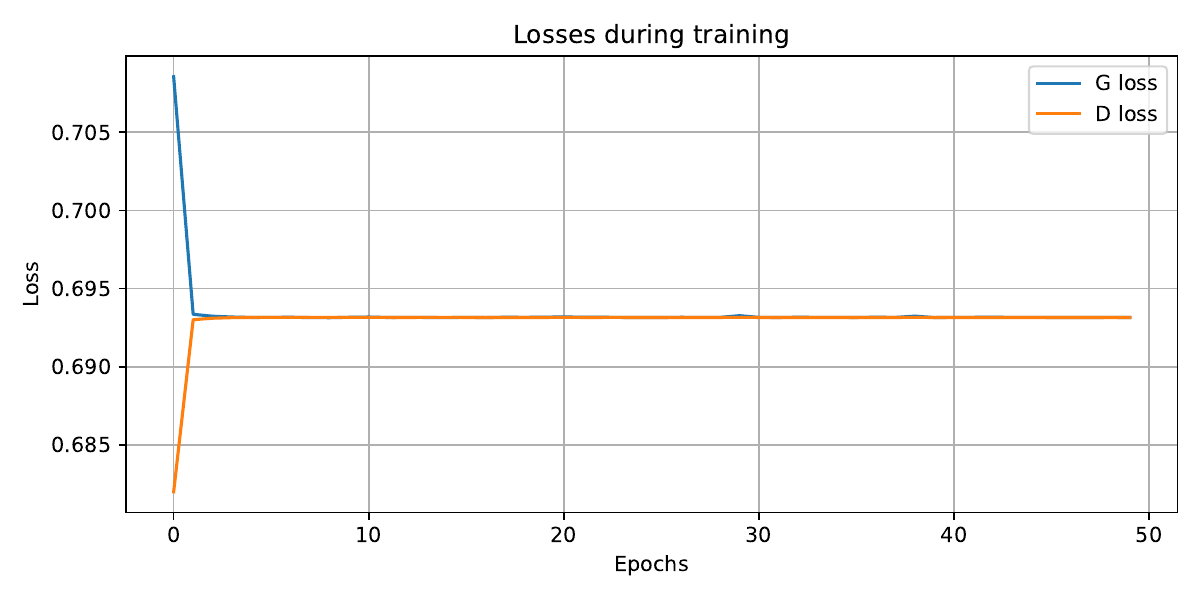}
\caption{ETH - data second period.}
\label{fig:lossethData2}
\end{subfigure}

\begin{subfigure}{0.55\textwidth}
\includegraphics[width=\textwidth]{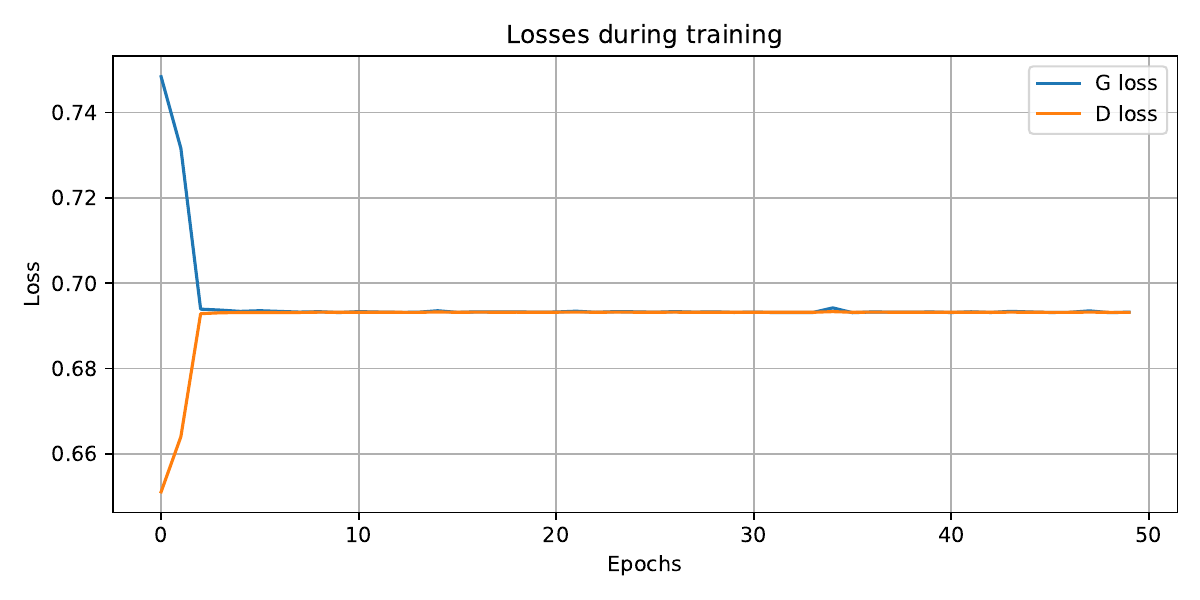}
\caption{ETH - data third period.}
\label{fig:lossethData3}
\end{subfigure}

\caption{Losses during training: ETH.}
\label{fig:lossethData}
\end{figure}

The Fig.~\ref{fig:lossethData} shows a comparison between the loss values of the generating and discriminating networks for the first, second, and third periods of ETH values.

\vspace*{-0.5cm}

\begin{figure}[H]
\centering

\begin{subfigure}{0.55\textwidth}
\includegraphics[width=\textwidth]{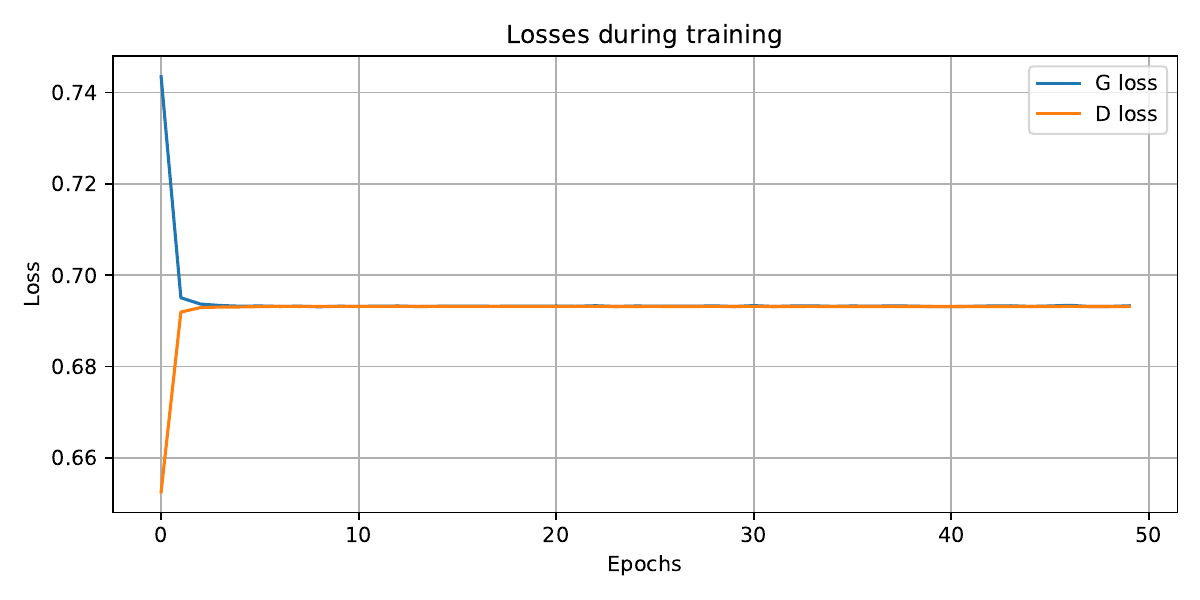}
\caption{XRP - data first period.}
\label{fig:lossxrpData1}
\end{subfigure}

\begin{subfigure}{0.55\textwidth}
\includegraphics[width=\textwidth]{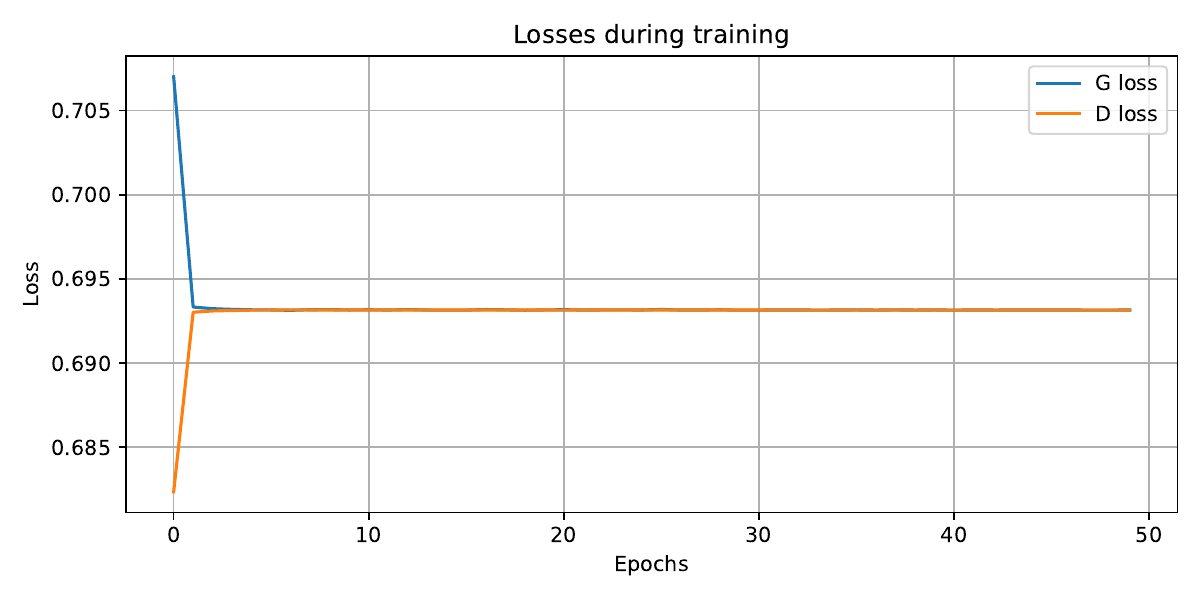}
\caption{XRP - data second period.}
\label{fig:lossxrpData2}
\end{subfigure}

\begin{subfigure}{0.55\textwidth}
\includegraphics[width=\textwidth]{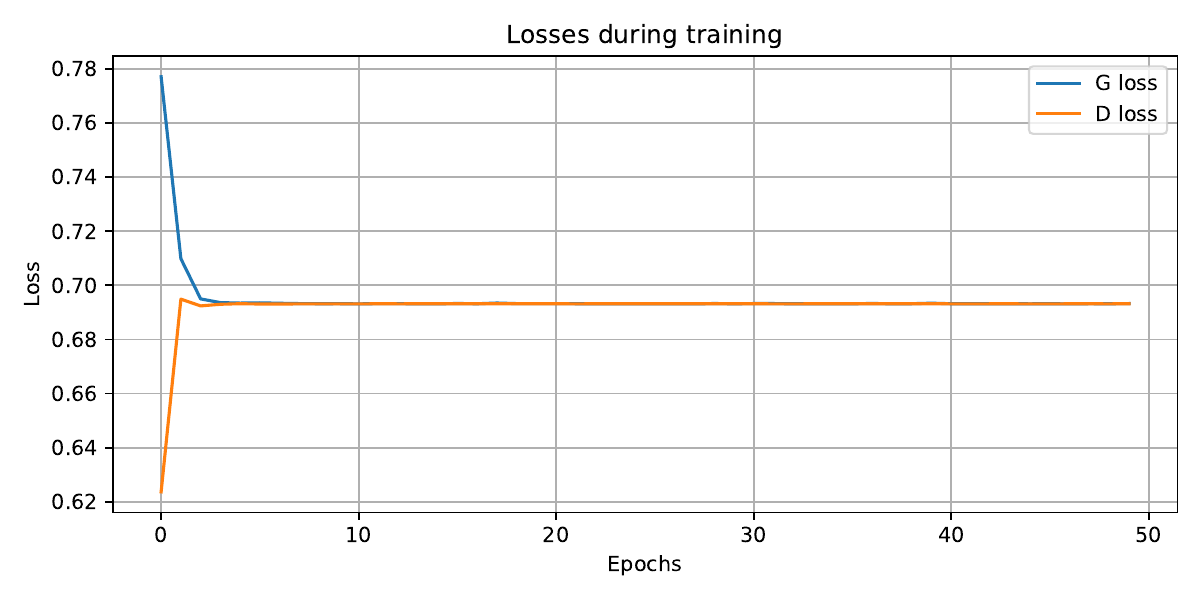}
\caption{XRP - data third period.}
\label{fig:lossxrpData3}
\end{subfigure}

\caption{Losses during training: XRP.}
\label{fig:lossxrpData}
\end{figure}

The Fig.~\ref{fig:lossxrpData} shows a comparison between the loss values of the generating and discriminating networks for the first, second, and third periods of XRP values.

\vspace*{-0.5cm}

\begin{figure}[H]
\centering

\begin{subfigure}{0.3\textwidth}
\includegraphics[width=\textwidth]{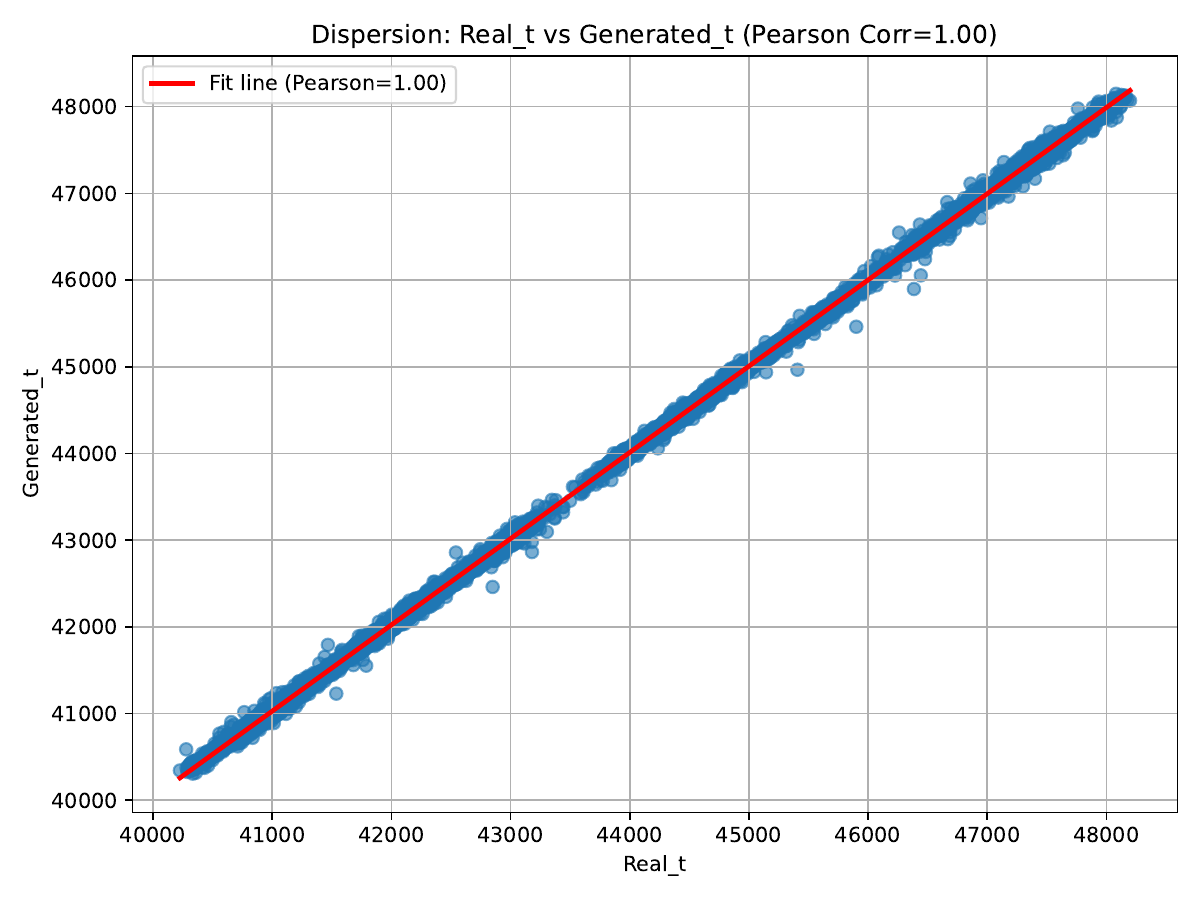}
\caption{BTC - data first period.}
\label{fig:disbtcdata1}
\end{subfigure}
\begin{subfigure}{0.3\textwidth}
\includegraphics[width=\textwidth]{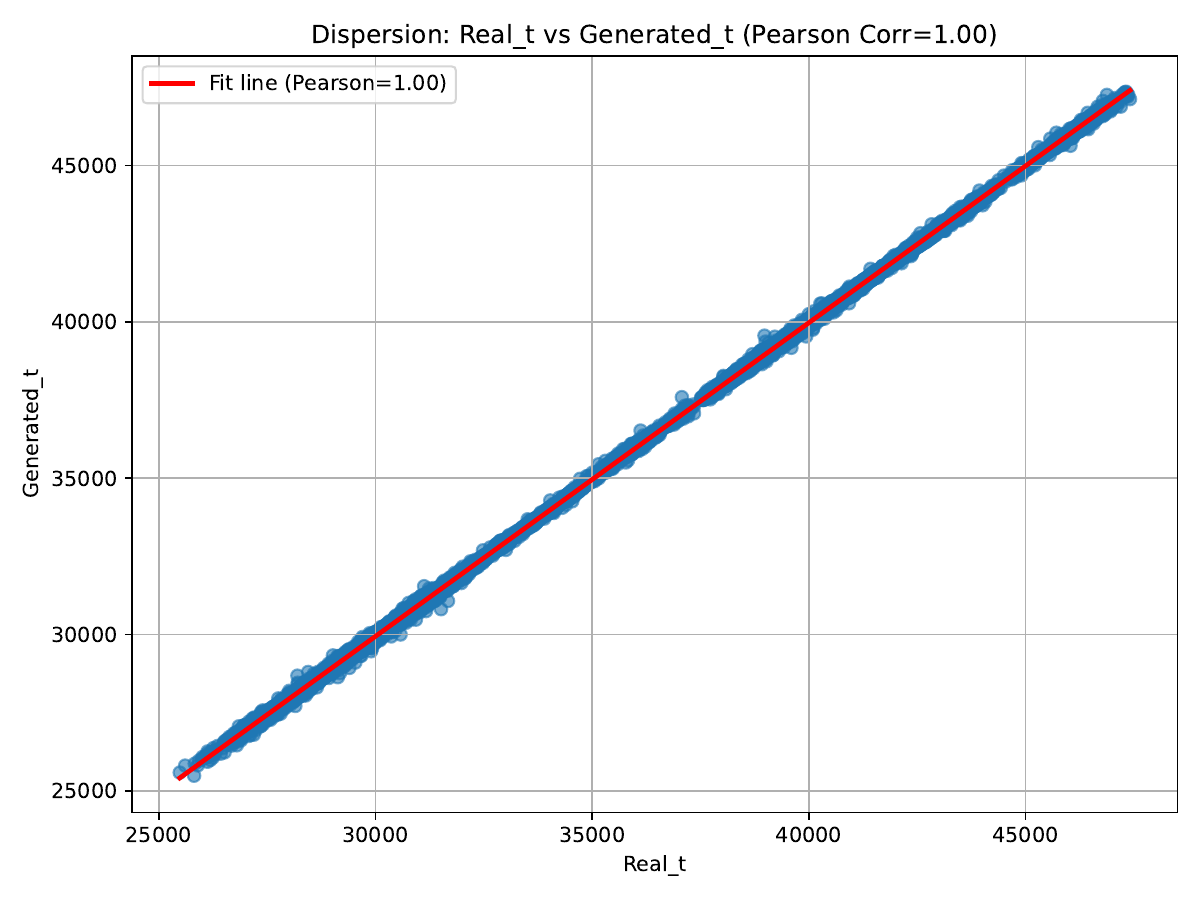}
\caption{BTC - data second period.}
\label{fig:disbtcdata2}
\end{subfigure}
\begin{subfigure}{0.3\textwidth}
\includegraphics[width=\textwidth]{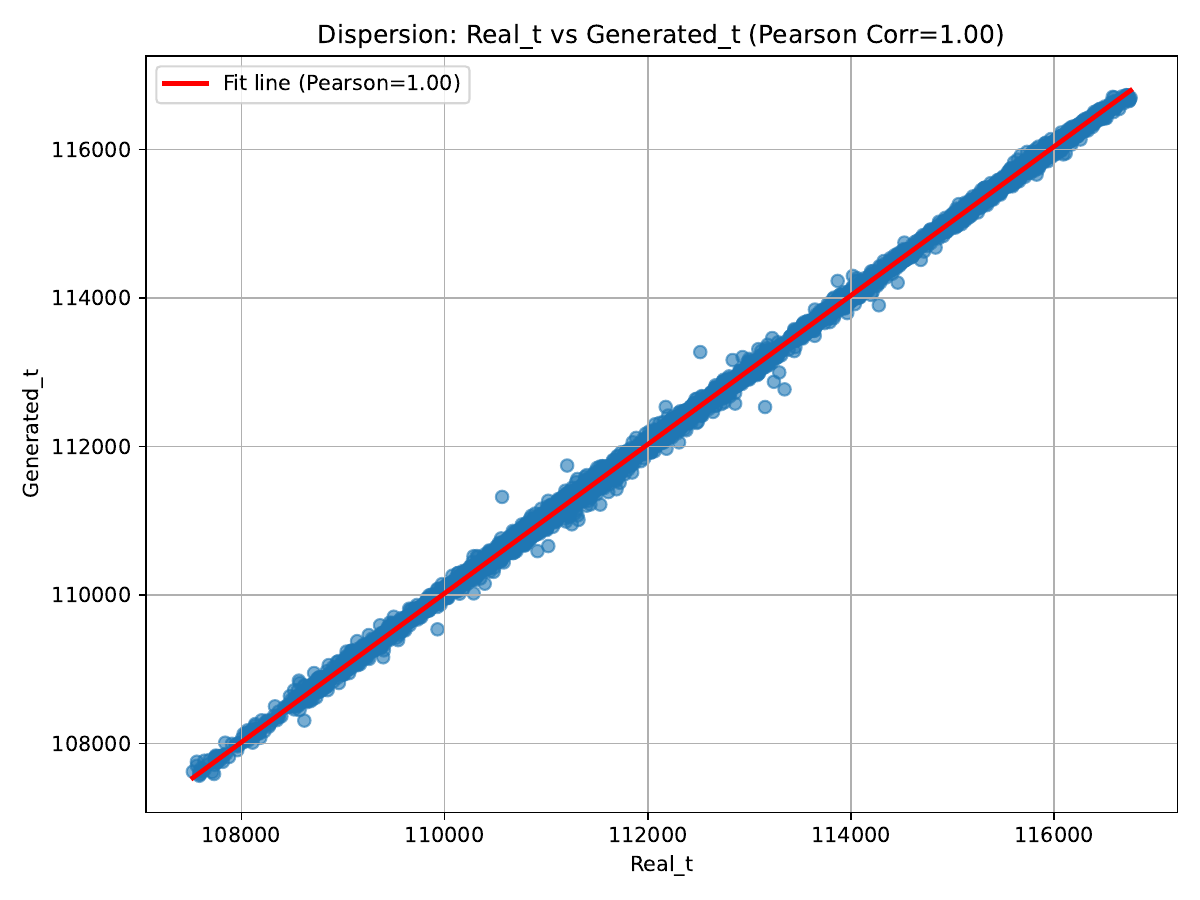}
\caption{BTC - data third period.}
\label{fig:disbtcdata3}
\end{subfigure}

\caption{Dispersion real vs generated - BTC.}
\label{fig:dispersionbtc}
\end{figure}

The Fig.~\ref{fig:dispersionbtc} shows the Pearson correlation between the true and generated values for the BTC data. The parameter results are: first period: Pearson - 0.9999 and Spearman - 0.9997; second period: Pearson - 1.0000 and Spearman - 0.9999; third period: Pearson - 0.9998 and Spearman - 0.9994. The MAE and RMSE values are: first period: MAE - 29.085452 and RMSE - 38.880509; second period: MAE - 46.711167 and RMSE - 58.489341; third period: MAE - 41.050079 and RMSE - 53.099261.

\vspace*{-0.5cm}

\begin{figure}[H]
\centering

\begin{subfigure}{0.3\textwidth}
\includegraphics[width=\textwidth]{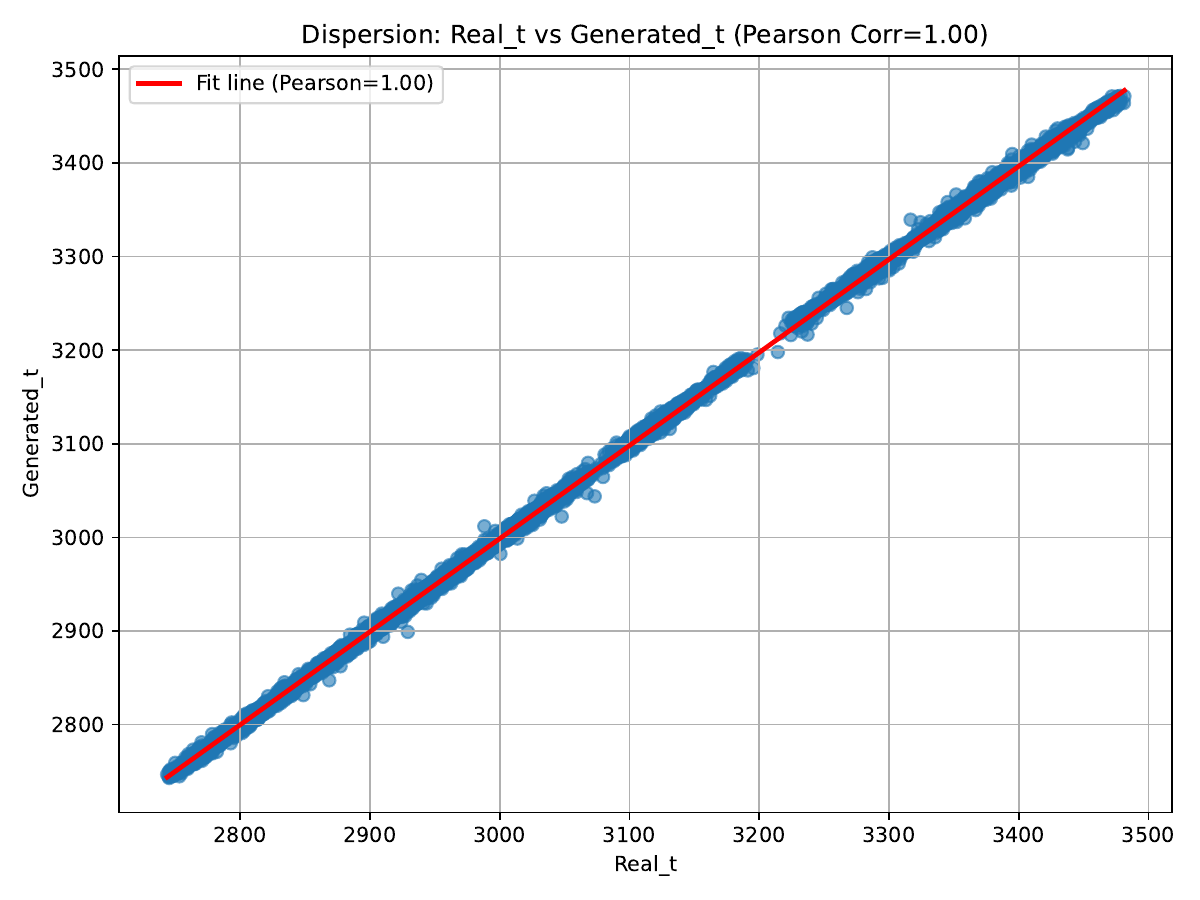}
\caption{ETH - data first period.}
\label{fig:disethdata1}
\end{subfigure}
\begin{subfigure}{0.3\textwidth}
\includegraphics[width=\textwidth]{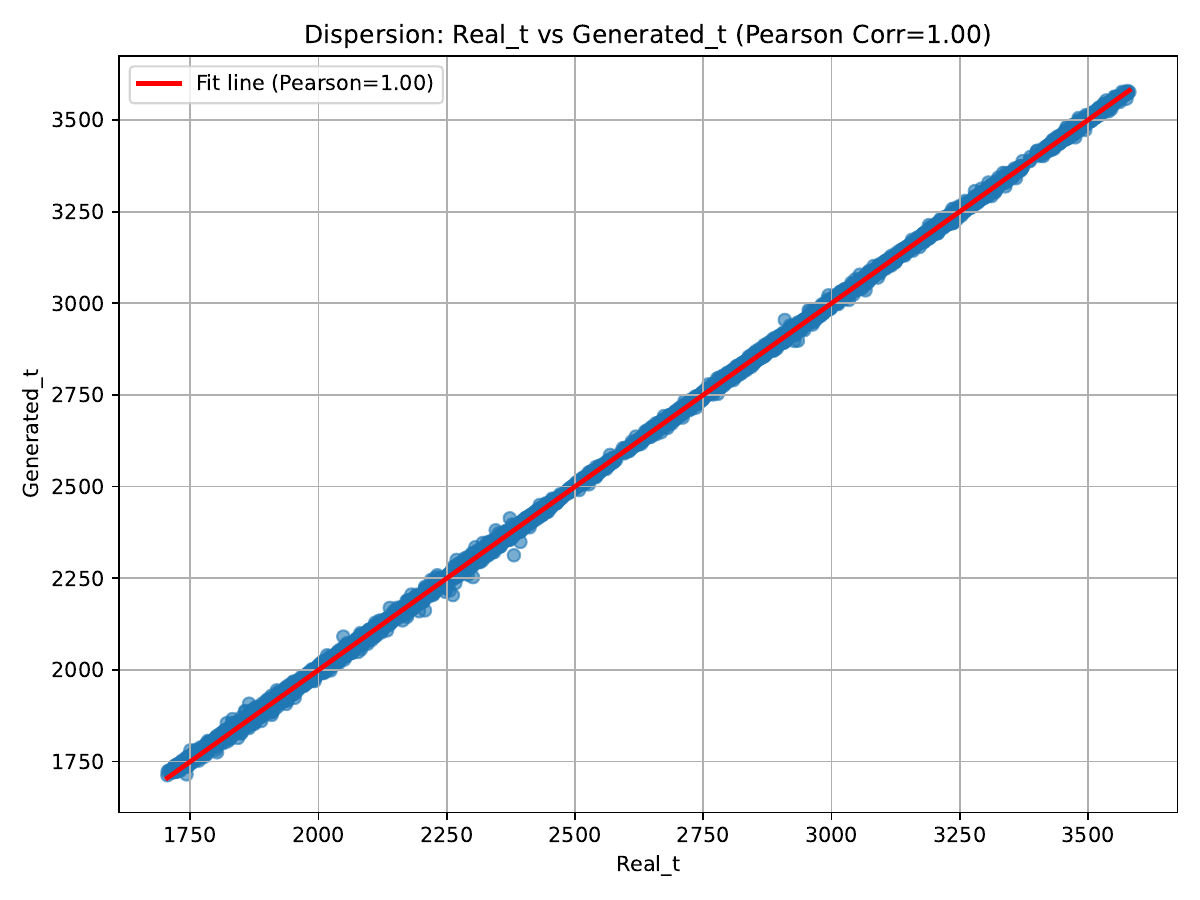}
\caption{ETH - data second period.}
\label{fig:disethdata2}
\end{subfigure}
\begin{subfigure}{0.3\textwidth}
\includegraphics[width=\textwidth]{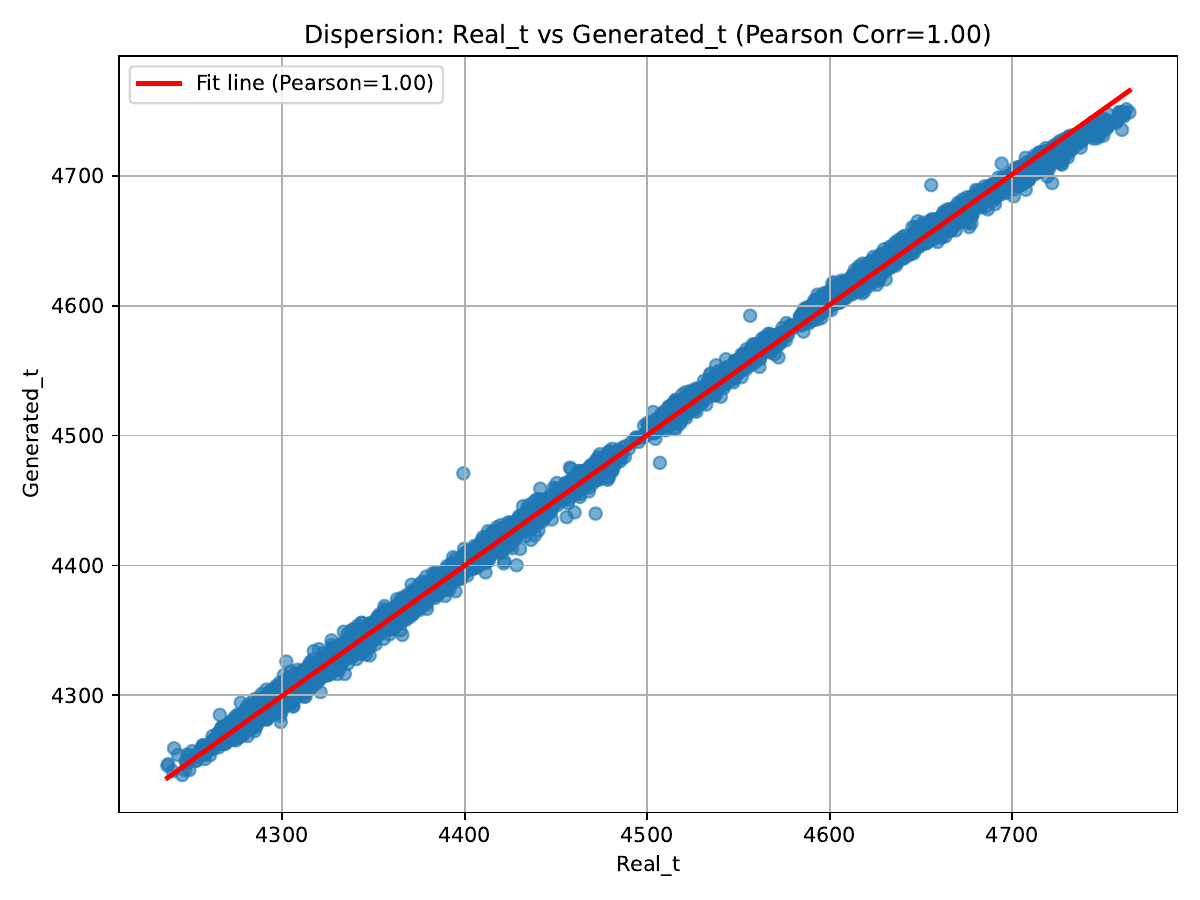}
\caption{ETH - data third period.}
\label{fig:disethdata3}
\end{subfigure}

\caption{Dispersion real vs generated - ETH.}
\label{fig:dispersioneth}
\end{figure}

The Fig.~\ref{fig:dispersioneth} shows the Pearson correlation between the true and generated values for the ETH data. The parameter results are: first period: Pearson - 0.9999 and Spearman - 0.9997; second period: Pearson - 1.0000 and Spearman - 0.9999; third period: Pearson - 0.9996 and Spearman - 0.9987. The MAE and RMSE values are: first period: MAE - 2.897669 and RMSE - 3.839272; second period: MAE - 2.495395 and RMSE - 3.541045; third period: MAE - 2.713196 and RMSE - 3.727772.

\vspace*{-0.5cm}

\begin{figure}[H]
\centering

\begin{subfigure}{0.3\textwidth}
\includegraphics[width=\textwidth]{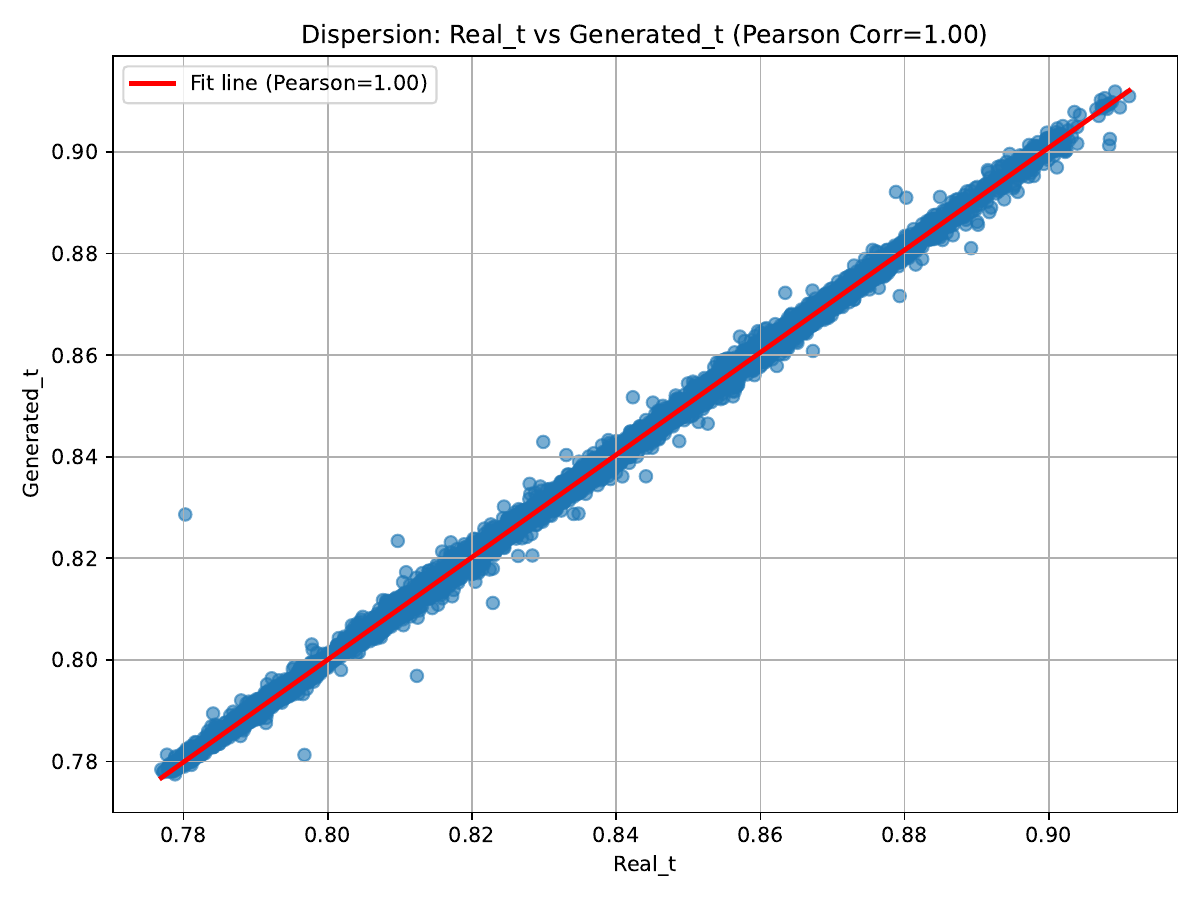}
\caption{XRP - data first period.}
\label{fig:disxrpdata1}
\end{subfigure}
\begin{subfigure}{0.3\textwidth}
\includegraphics[width=\textwidth]{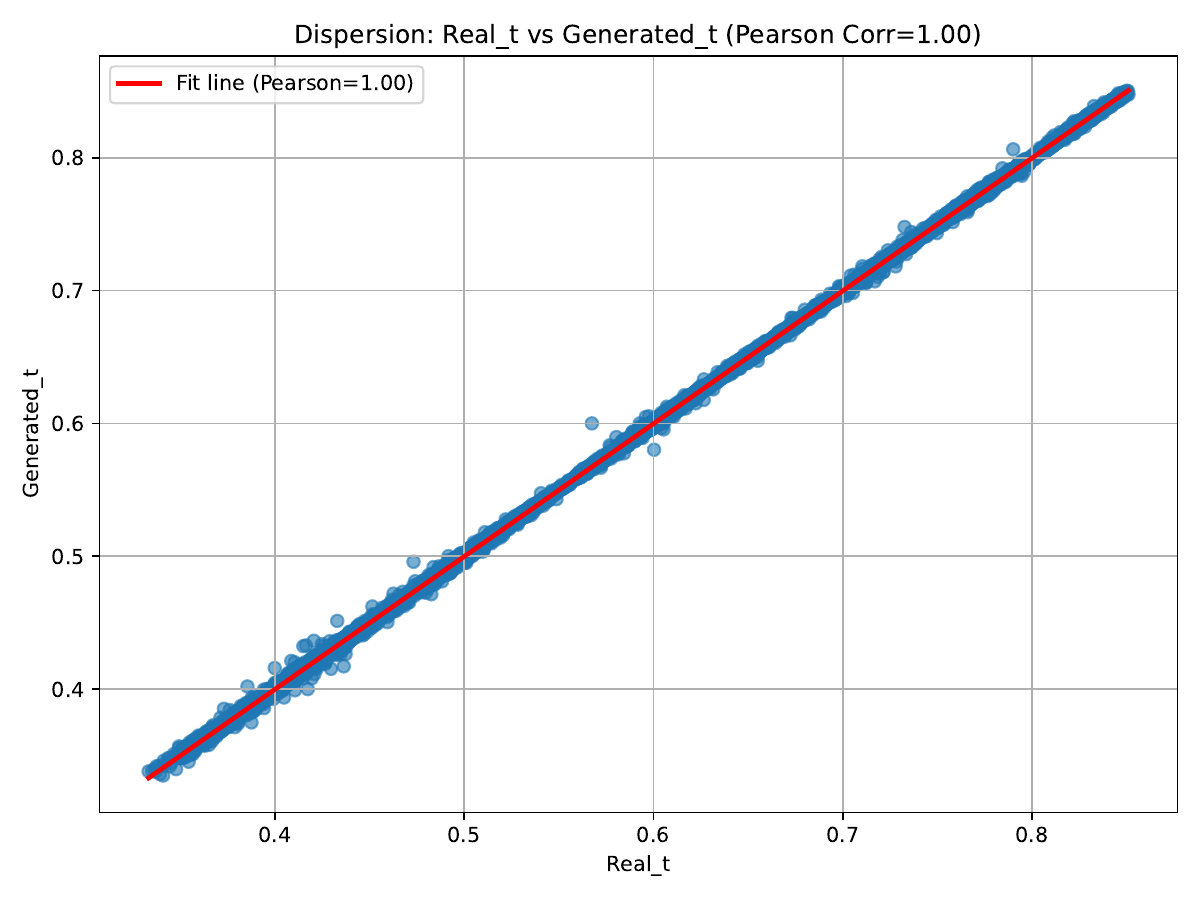}
\caption{XRP - data second period.}
\label{fig:disxrpdata2}
\end{subfigure}
\begin{subfigure}{0.3\textwidth}
\includegraphics[width=\textwidth]{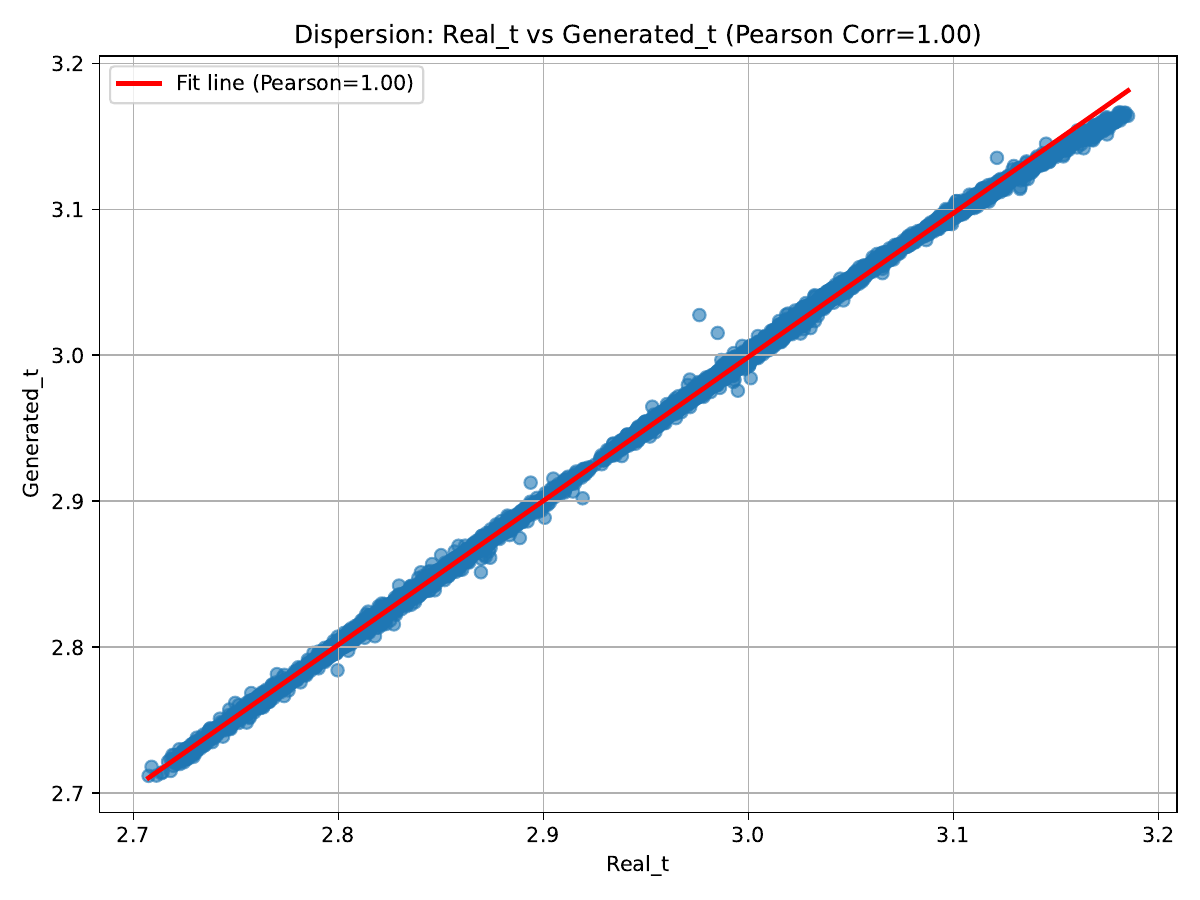}
\caption{XRP - data third period.}
\label{fig:disxrpdata3}
\end{subfigure}

\caption{Dispersion real vs generated - XRP.}
\label{fig:dispersionxrp}
\end{figure}

The Fig.~\ref{fig:dispersionxrp} shows the Pearson correlation between the true and generated values for the XRP data. The parameter results are: first period: Pearson - 0.9994 and Spearman - 0.9993; second period: Pearson - 1.0000 and Spearman - 0.9999; third period: Pearson - 0.9997 and Spearman - 0.9997. The MAE and RMSE values are: first period: MAE - 0.000683 and RMSE - 0.001018; second period: MAE - 0.000656 and RMSE - 0.000945; third period: MAE - 0.001976 and RMSE - 0.003112.

\vspace*{-0.5cm}

\begin{figure}[H]
\centering

\begin{subfigure}{0.65\textwidth}
\includegraphics[width=\textwidth]{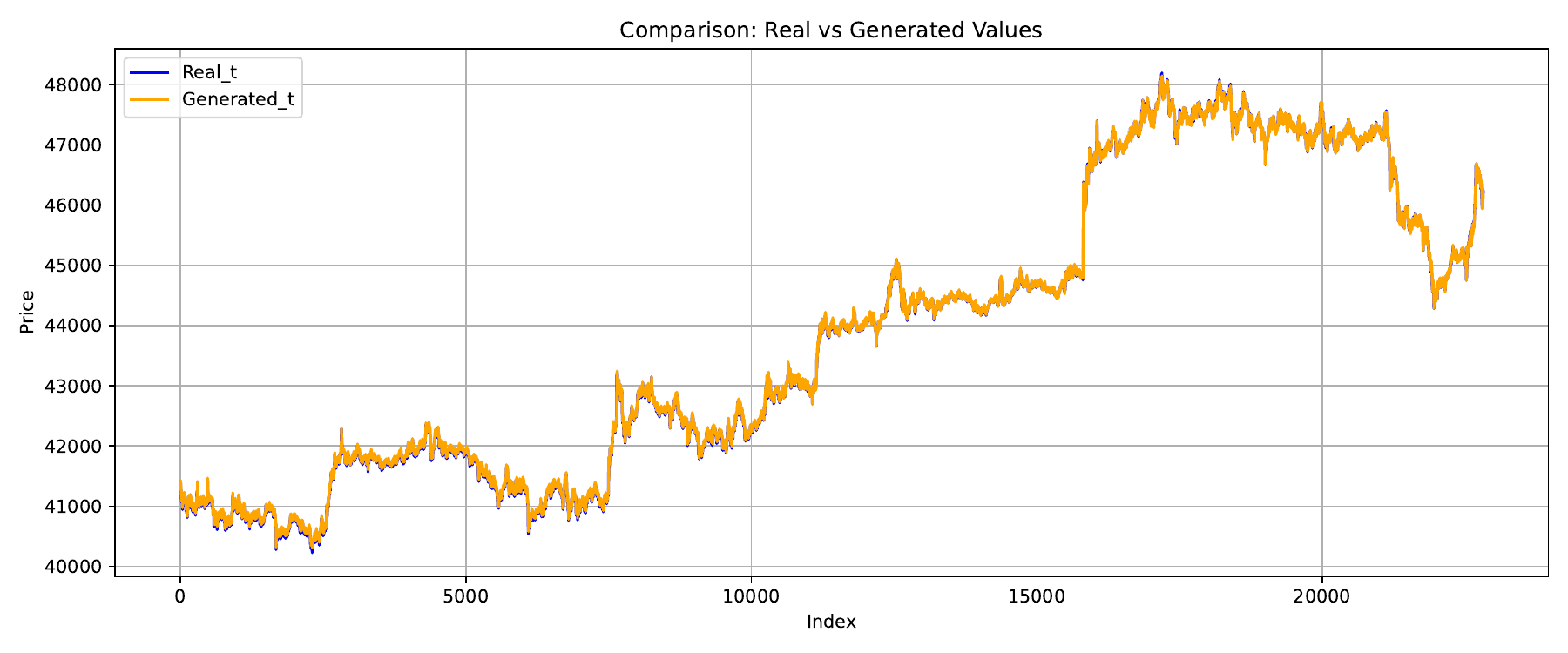}
\caption{BTC - data first period.}
\label{fig:BTCdata1}
\end{subfigure}

\begin{subfigure}{0.65\textwidth}
\includegraphics[width=\textwidth]{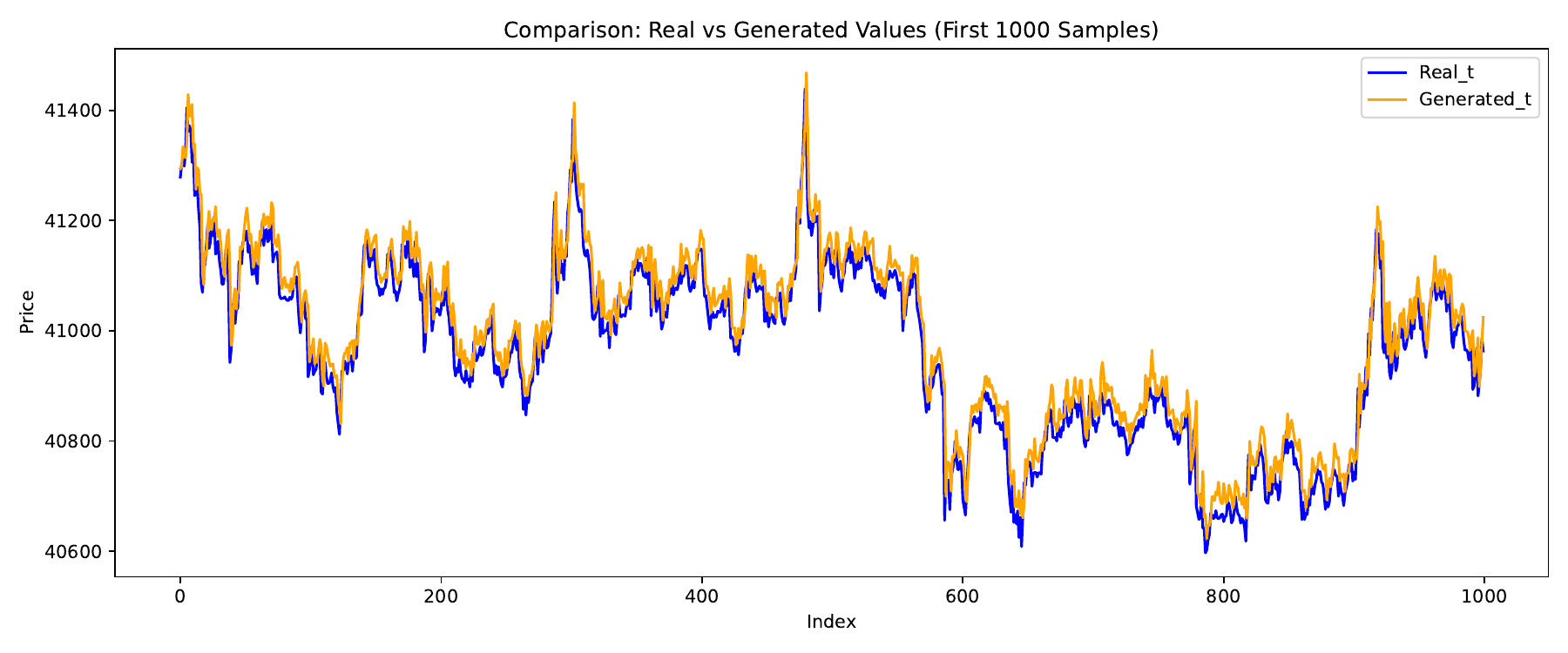}
\caption{BTC - data first period - first 1000 samples.}
\label{fig:BTCdata1-1000samples}
\end{subfigure}

\caption{Original series vs generated series: BTC - data first period.}
\label{fig:BTCdata1-result}
\end{figure}

The Fig.~\ref{fig:BTCdata1-result} shows a comparison between the generated and actual values for the first period of BTC values, and a sample of the first 1000 samples for better visualization.

\vspace*{-0.5cm}

\begin{figure}[H]
\centering

\begin{subfigure}{0.65\textwidth}
\includegraphics[width=\textwidth]{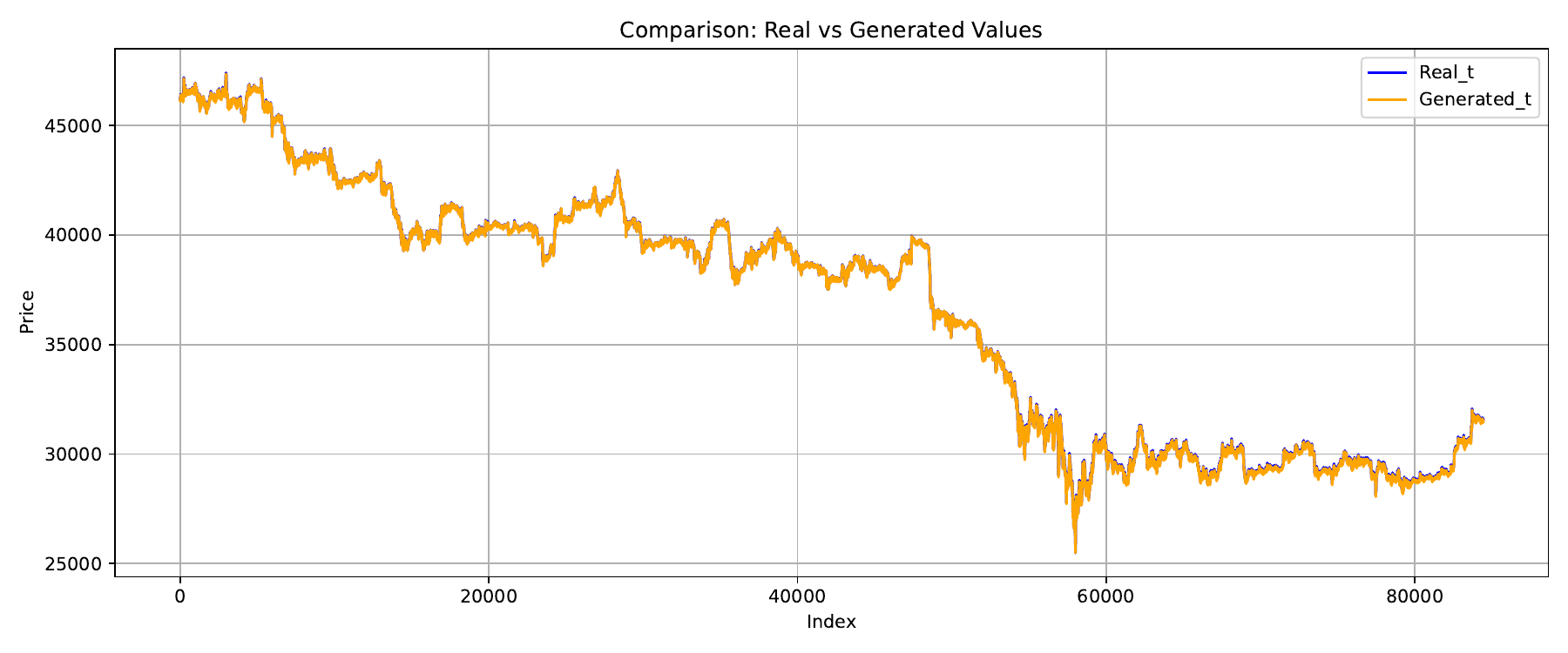}
\caption{BTC - data second period.}
\label{fig:BTCdata2}
\end{subfigure}

\begin{subfigure}{0.65\textwidth}
\includegraphics[width=\textwidth]{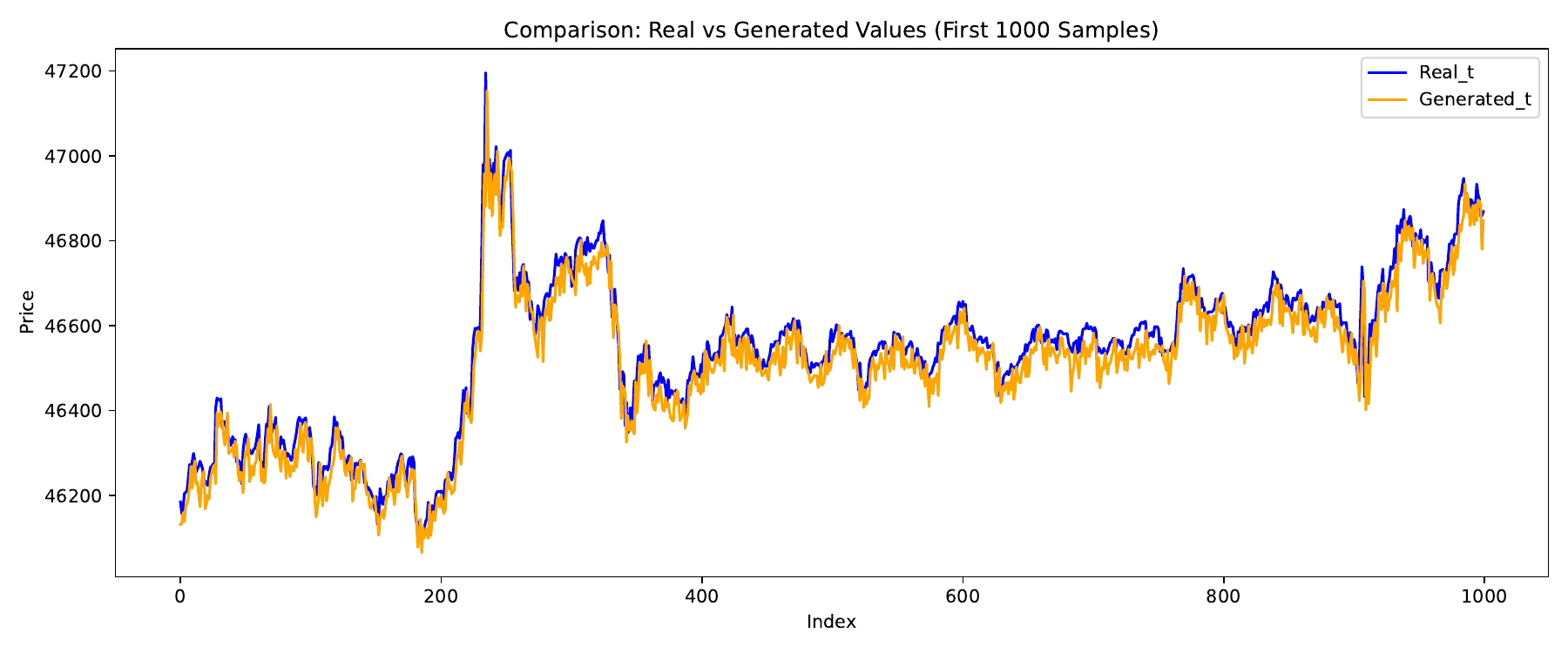}
\caption{BTC - data second period - first 1000 samples.}
\label{fig:BTCdata2-1000samples}
\end{subfigure}

\caption{Original series vs generated series: BTC - data second period.}
\label{fig:BTCdata2-result}
\end{figure}

The Fig.~\ref{fig:BTCdata2-result} shows a comparison between the generated and actual values for the second period of BTC values, and a sample of the first 1000 samples for better visualization.

\vspace*{-0.5cm}

\begin{figure}[H]
\centering

\begin{subfigure}{0.65\textwidth}
\includegraphics[width=\textwidth]{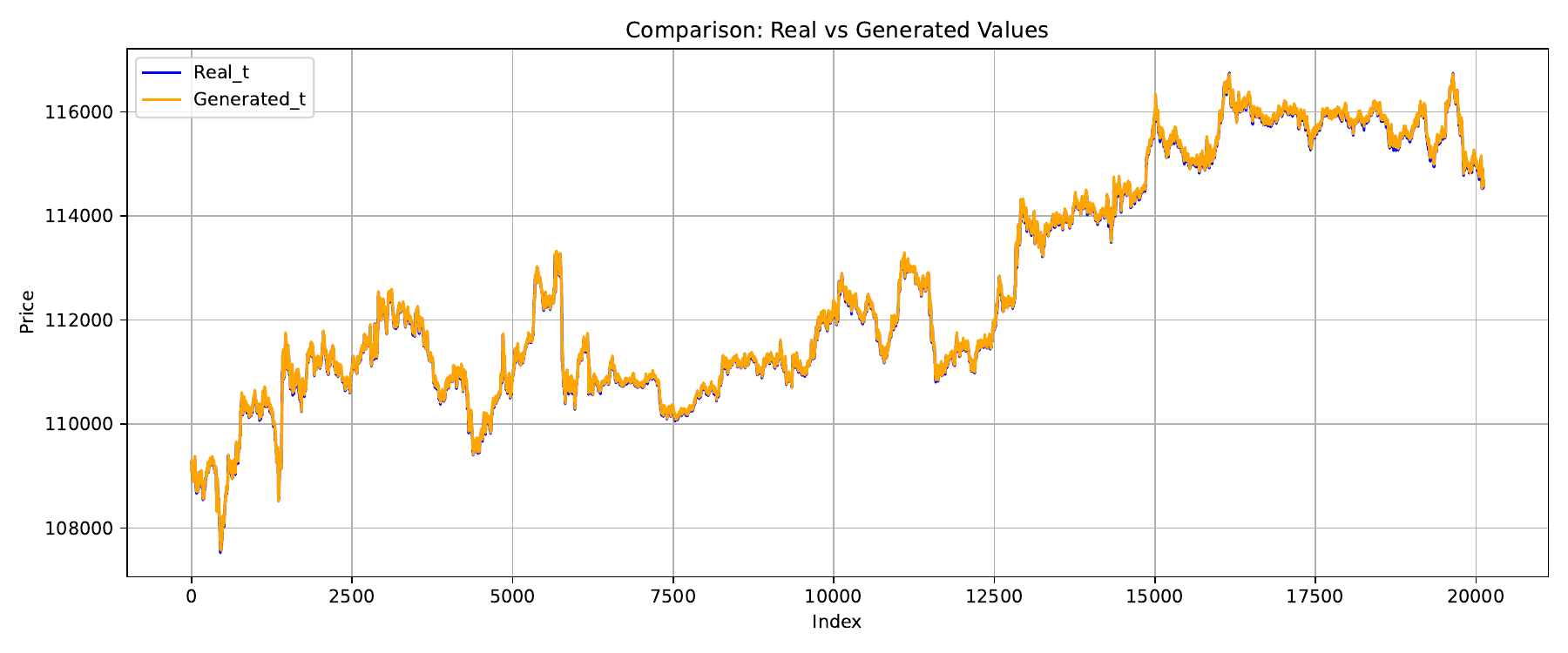}
\caption{BTC - data third period.}
\label{fig:BTCdata3}
\end{subfigure}

\begin{subfigure}{0.65\textwidth}
\includegraphics[width=\textwidth]{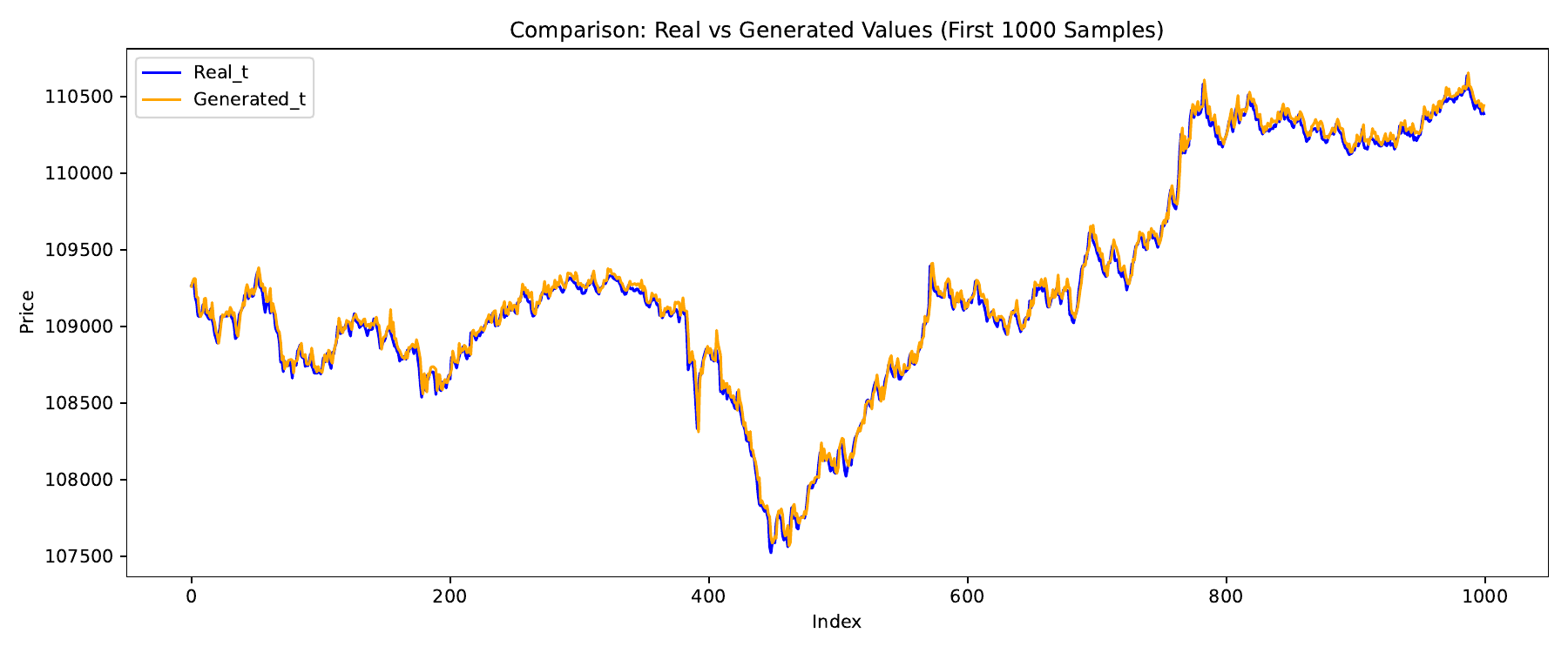}
\caption{BTC - data third period - first 1000 samples.}
\label{fig:BTCdata3-1000samples}
\end{subfigure}

\caption{Original series vs generated series: BTC - data third period.}
\label{fig:BTCdata3-result}
\end{figure}

The Fig.~\ref{fig:BTCdata3-result} shows a comparison between the generated and actual values for the third period of BTC values, and a sample of the first 1000 samples for better visualization.

\vspace*{-0.5cm}

\begin{figure}[H]
\centering

\begin{subfigure}{0.65\textwidth}
\includegraphics[width=\textwidth]{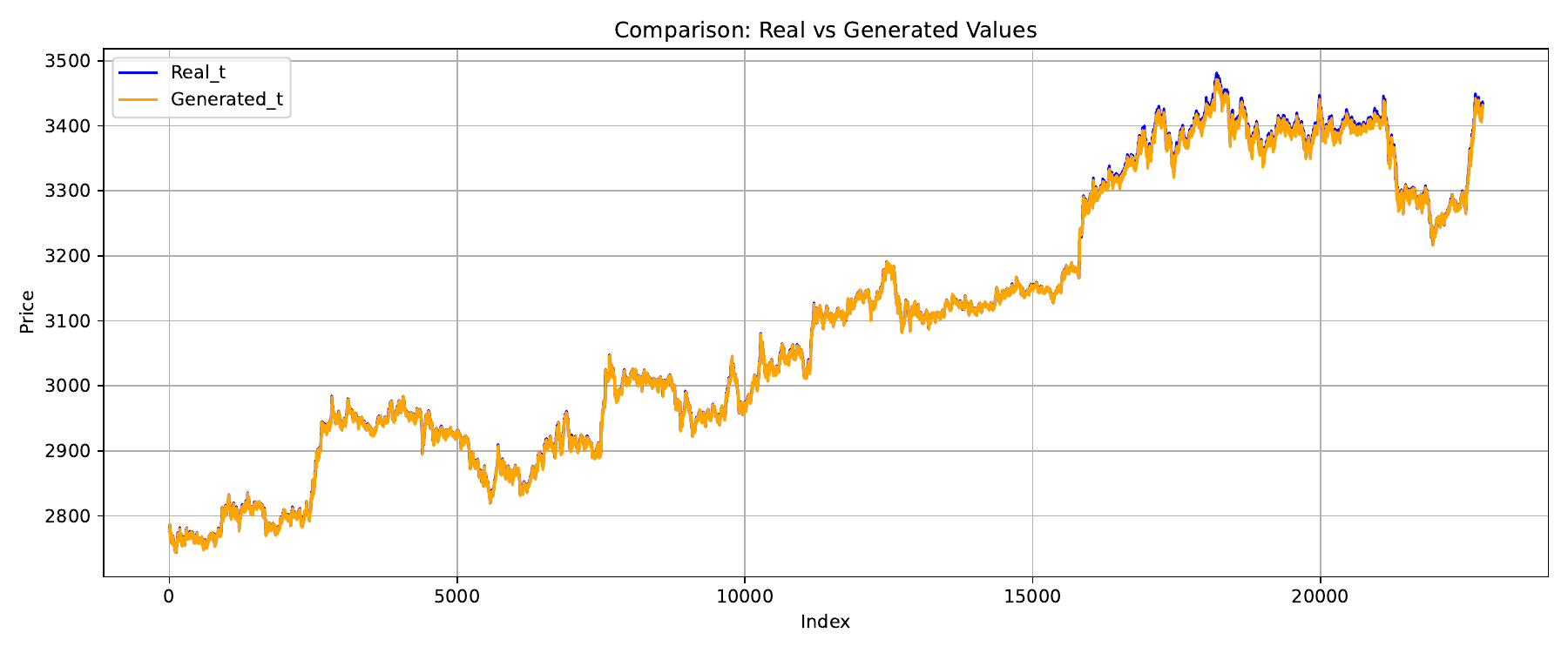}
\caption{ETH - data first period.}
\label{fig:ETHdata1}
\end{subfigure}

\begin{subfigure}{0.65\textwidth}
\includegraphics[width=\textwidth]{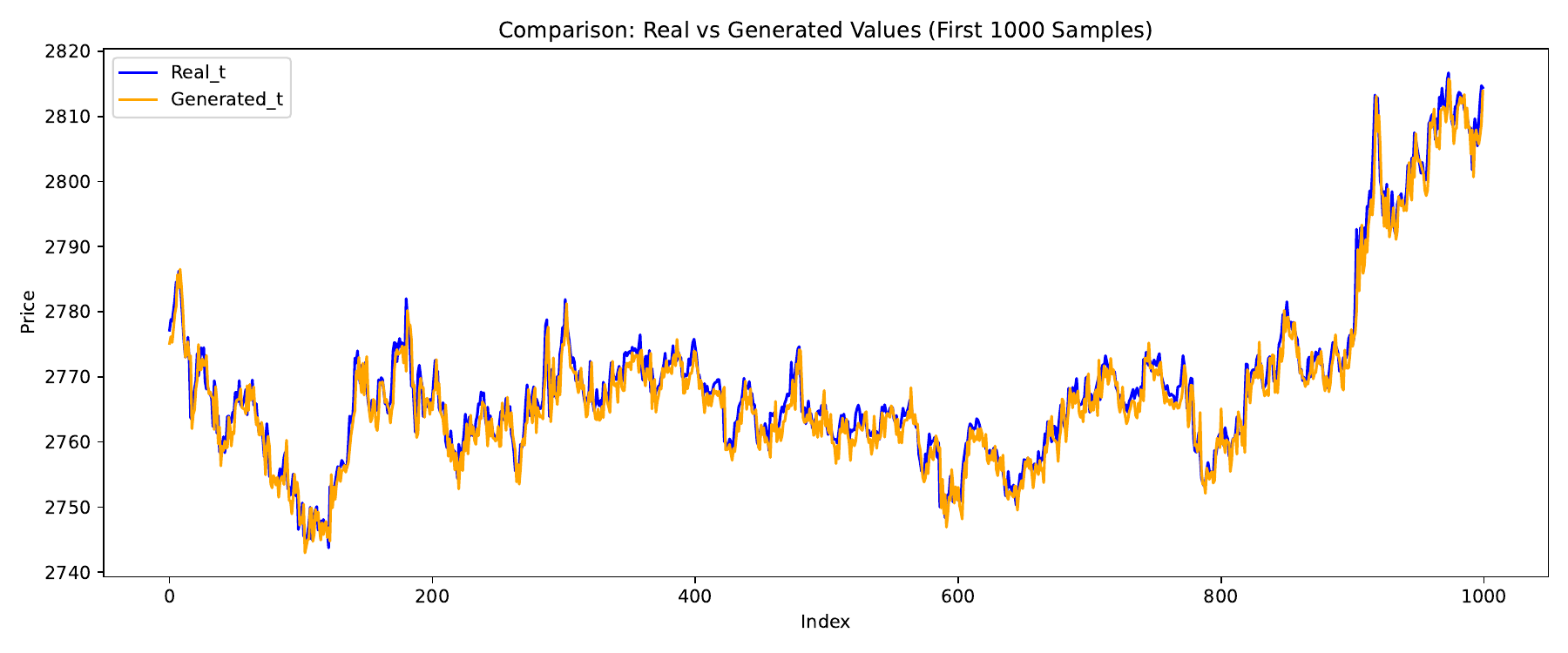}
\caption{ETH - data first period - first 1000 samples.}
\label{fig:ETHdata1-1000samples}
\end{subfigure}

\caption{Original series vs generated series: ETH - data first period.}
\label{fig:ETHdata1-result}
\end{figure}

The Fig.~\ref{fig:ETHdata1-result} shows a comparison between the generated and actual values for the first period of ETH values, and a sample of the first 1000 samples for better visualization.

\vspace*{-0.5cm}

\begin{figure}[H]
\centering

\begin{subfigure}{0.65\textwidth}
\includegraphics[width=\textwidth]{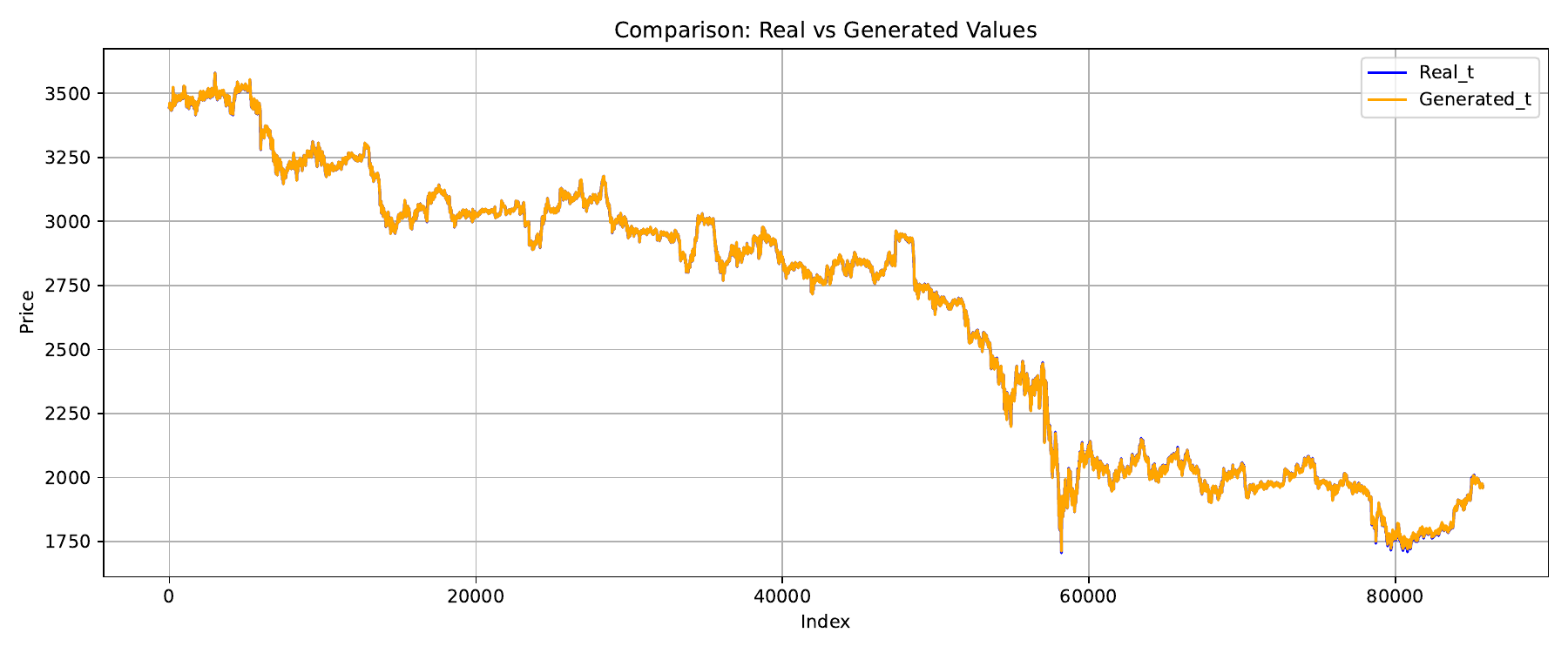}
\caption{ETH - data second period.}
\label{fig:ETHdata2}
\end{subfigure}

\begin{subfigure}{0.65\textwidth}
\includegraphics[width=\textwidth]{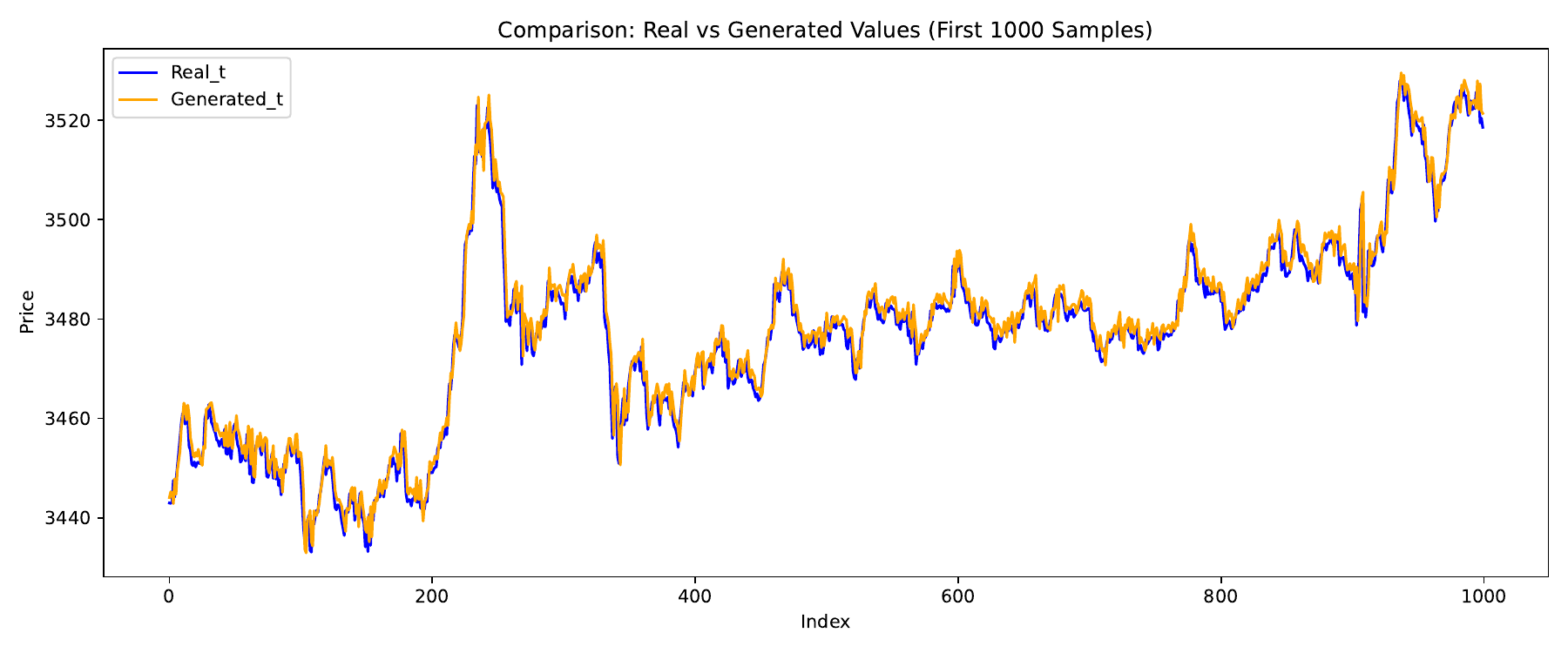}
\caption{ETH - data second period - first 1000 samples.}
\label{fig:ETHdata2-1000samples}
\end{subfigure}

\caption{Original series vs generated series: ETH - data second period.}
\label{fig:ETHdata2-result}
\end{figure}

The Fig.~\ref{fig:ETHdata2-result} shows a comparison between the generated and actual values for the second period of ETH values, and a sample of the first 1000 samples for better visualization.

\vspace*{-0.5cm}

\begin{figure}[H]
\centering

\begin{subfigure}{0.65\textwidth}
\includegraphics[width=\textwidth]{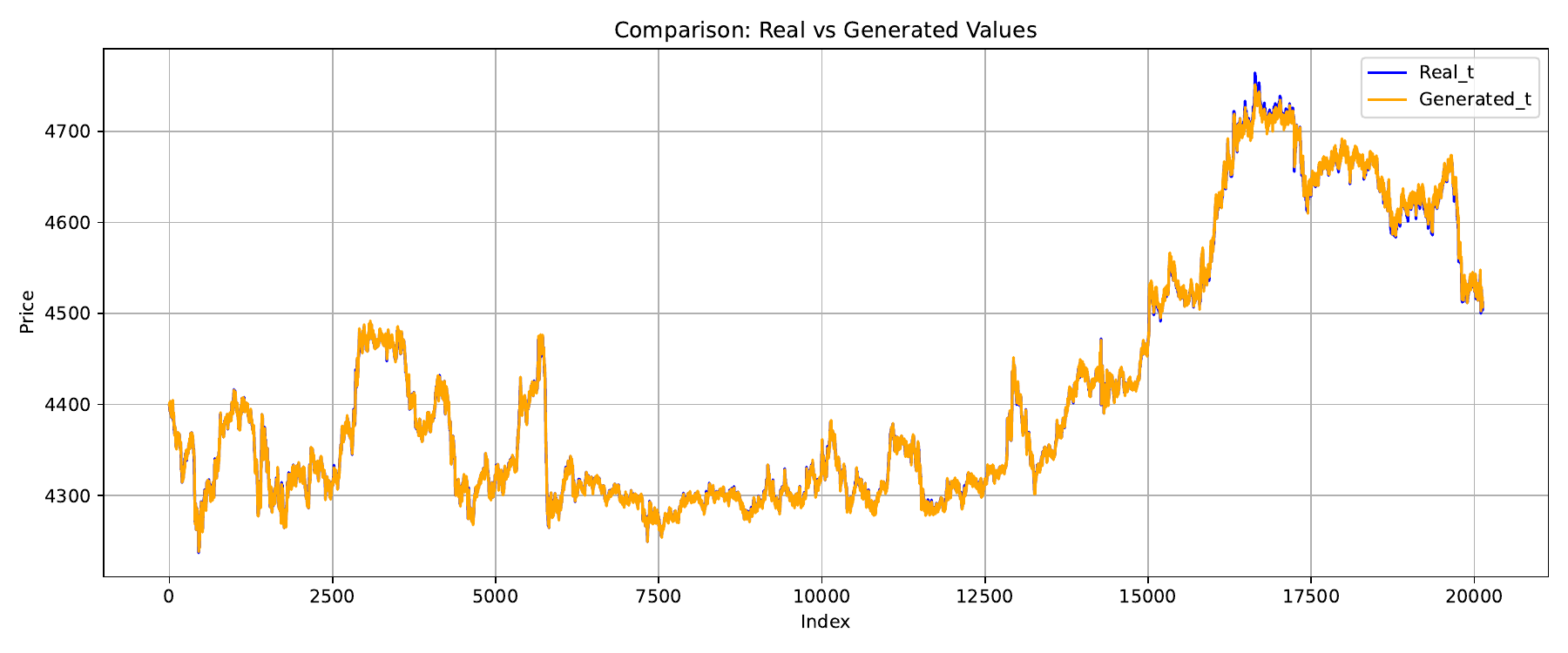}
\caption{ETH - data third period.}
\label{fig:ETHdata3}
\end{subfigure}

\begin{subfigure}{0.65\textwidth}
\includegraphics[width=\textwidth]{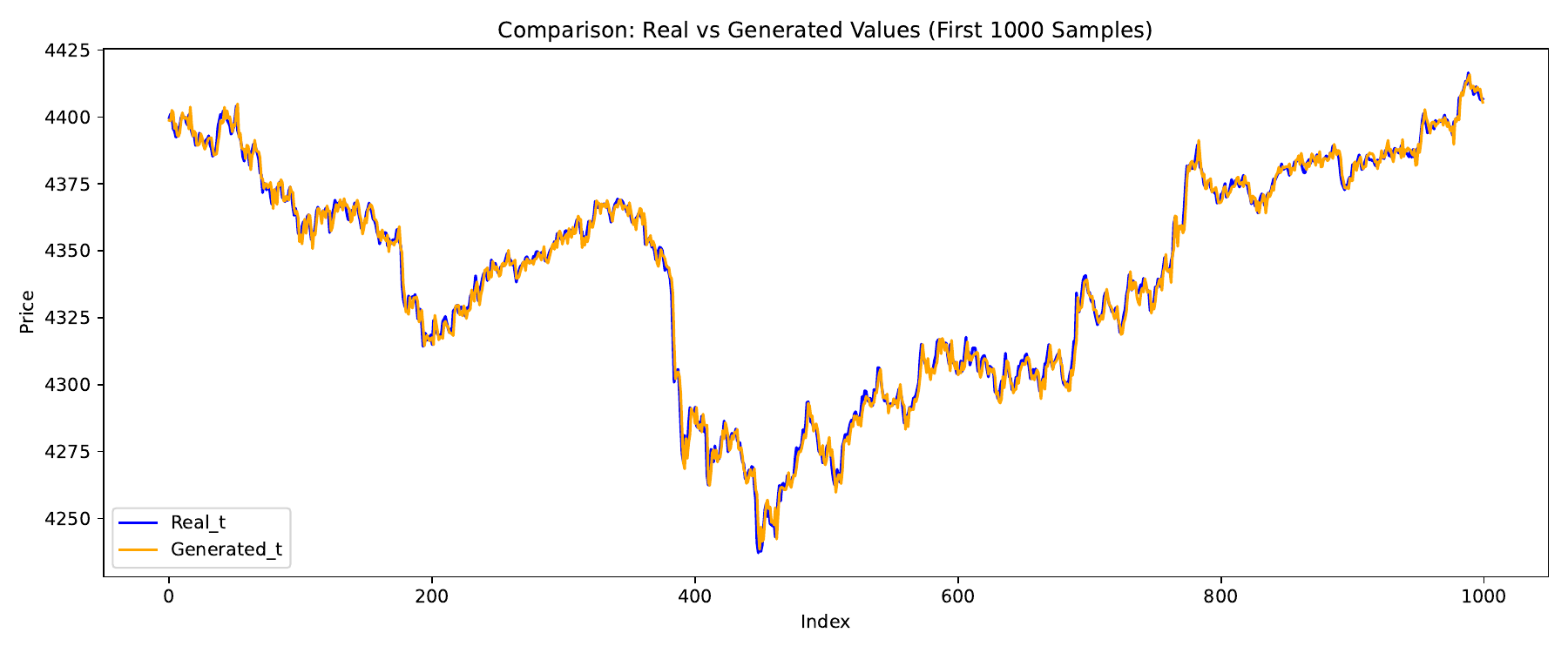}
\caption{ETH - data third period - first 1000 samples.}
\label{fig:ETHdata3-1000samples}
\end{subfigure}

\caption{Original series vs generated series: ETH - data third period.}
\label{fig:ETHdata3-result}
\end{figure}

The Fig.~\ref{fig:ETHdata3-result} shows a comparison between the generated and actual values for the third period of ETH values, and a sample of the first 1000 samples for better visualization.

\vspace*{-0.5cm}

\begin{figure}[H]
\centering

\begin{subfigure}{0.65\textwidth}
\includegraphics[width=\textwidth]{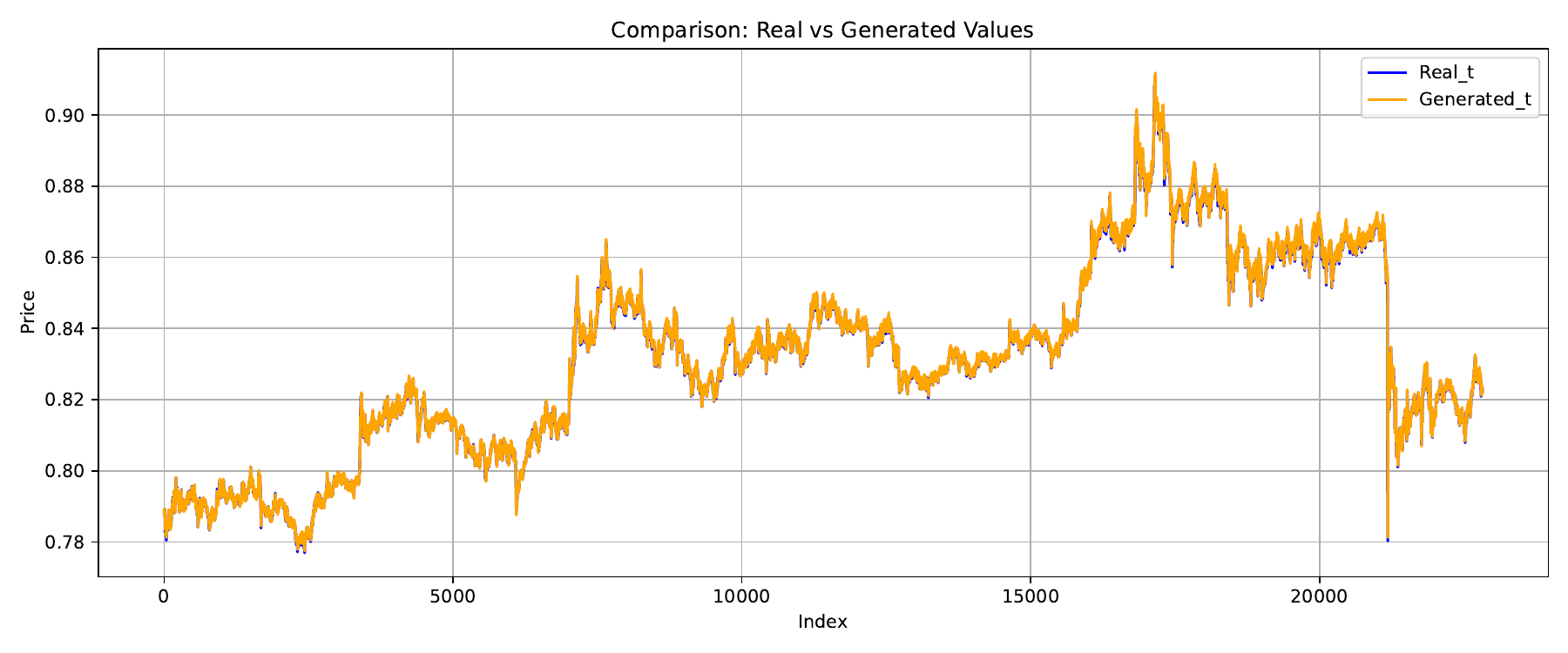}
\caption{XRP - data first period.}
\label{fig:XRPdata1}
\end{subfigure}

\begin{subfigure}{0.65\textwidth}
\includegraphics[width=\textwidth]{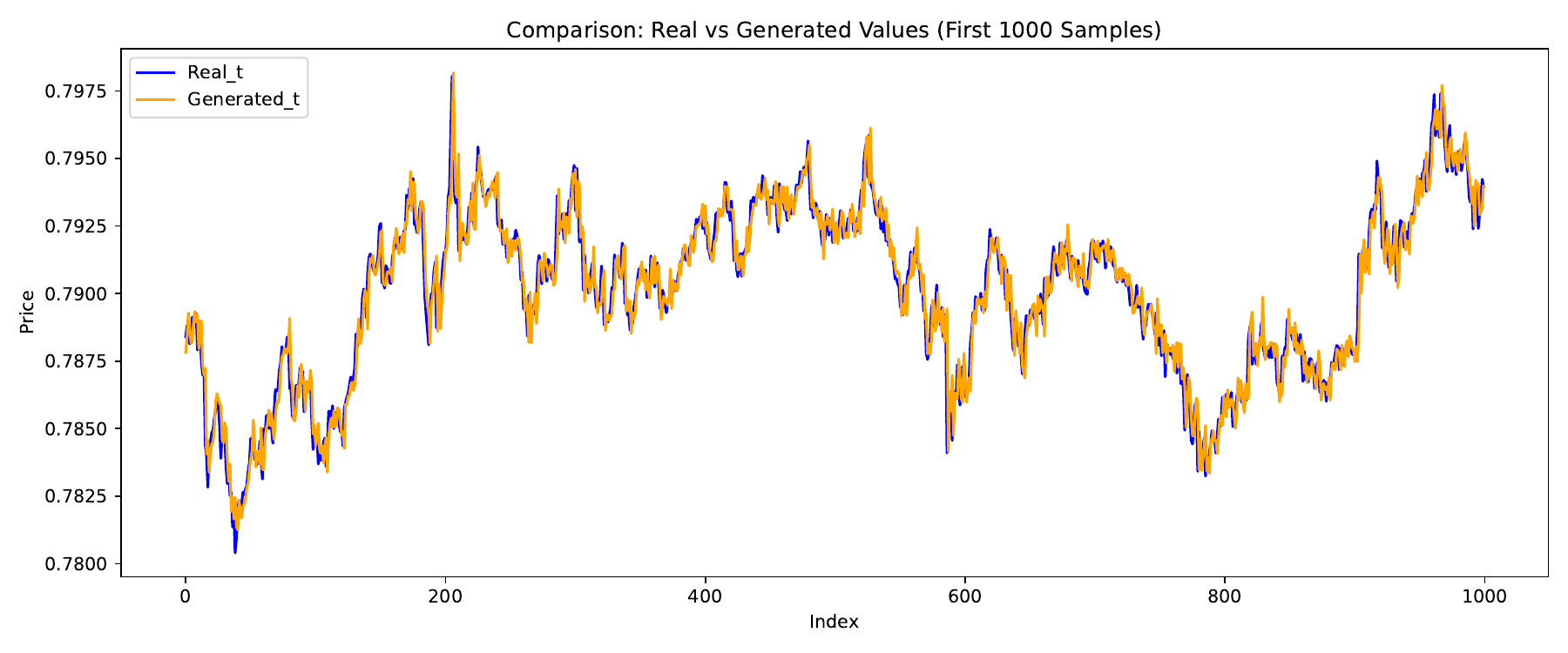}
\caption{XRP - data first period - first 1000 samples.}
\label{fig:XRPdata1-1000samples}
\end{subfigure}

\caption{Original series vs generated series: XRP - data first period.}
\label{fig:XRPdata1-result}
\end{figure}

The Fig.~\ref{fig:XRPdata1-result} shows a comparison between the generated and actual values for the first period of XRP values, and a sample of the first 1000 samples for better visualization.

\vspace*{-0.5cm}

\begin{figure}[H]
\centering

\begin{subfigure}{0.65\textwidth}
\includegraphics[width=\textwidth]{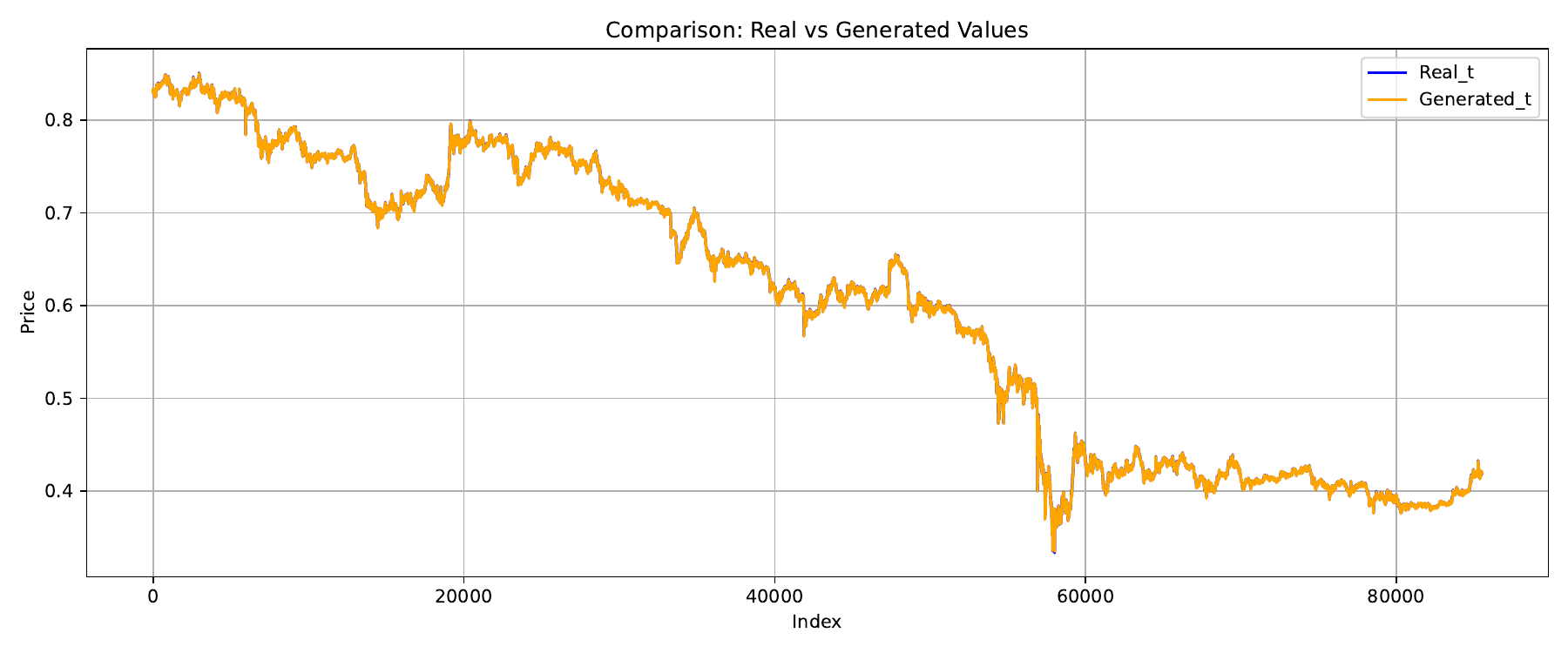}
\caption{XRP - data second period.}
\label{fig:XRPdata2}
\end{subfigure}

\begin{subfigure}{0.65\textwidth}
\includegraphics[width=\textwidth]{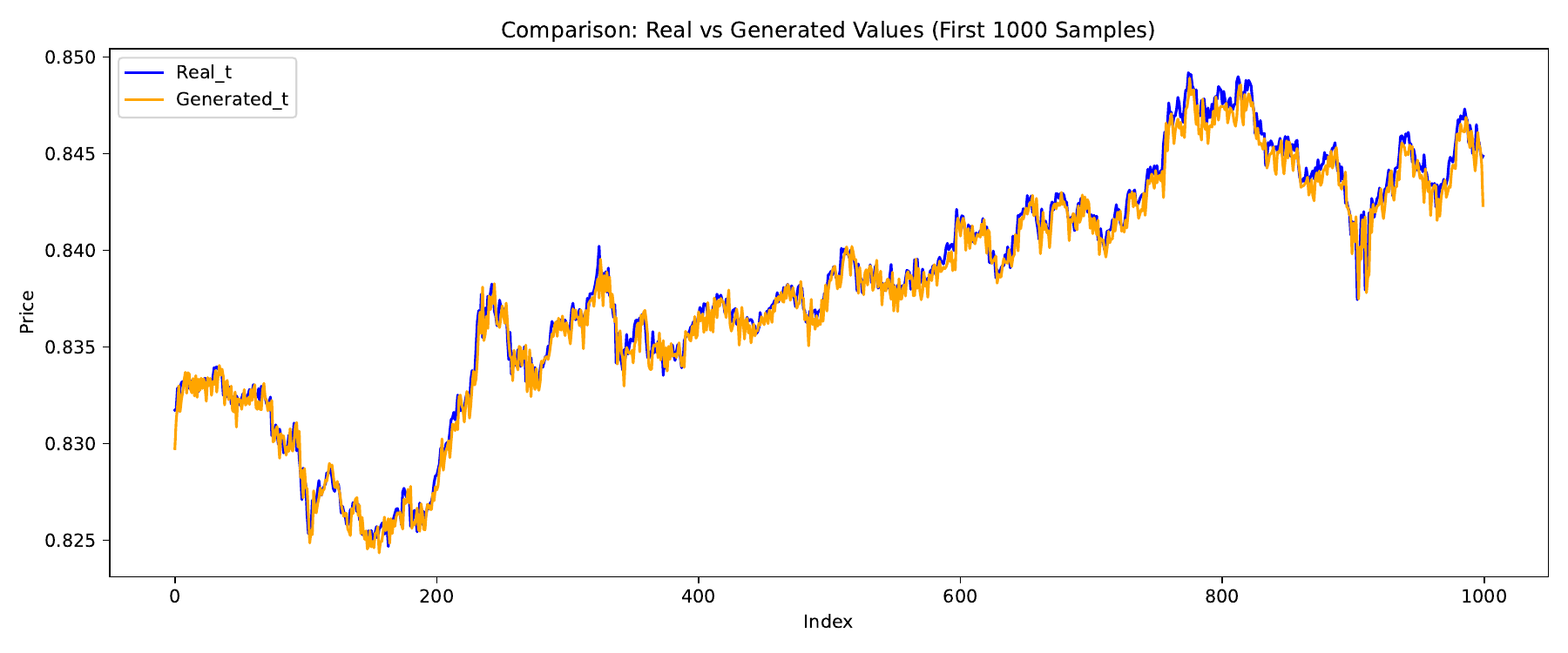}
\caption{XRP - data second period - first 1000 samples.}
\label{fig:XRPdata2-1000samples}
\end{subfigure}

\caption{Original series vs generated series: XRP - data second period.}
\label{fig:XRPdata2-result}
\end{figure}

The Fig.~\ref{fig:XRPdata2-result} shows a comparison between the generated and actual values for the second period of XRP values, and a sample of the first 1000 samples for better visualization.

\vspace*{-0.5cm}

\begin{figure}[H]
\centering

\begin{subfigure}{0.65\textwidth}
\includegraphics[width=\textwidth]{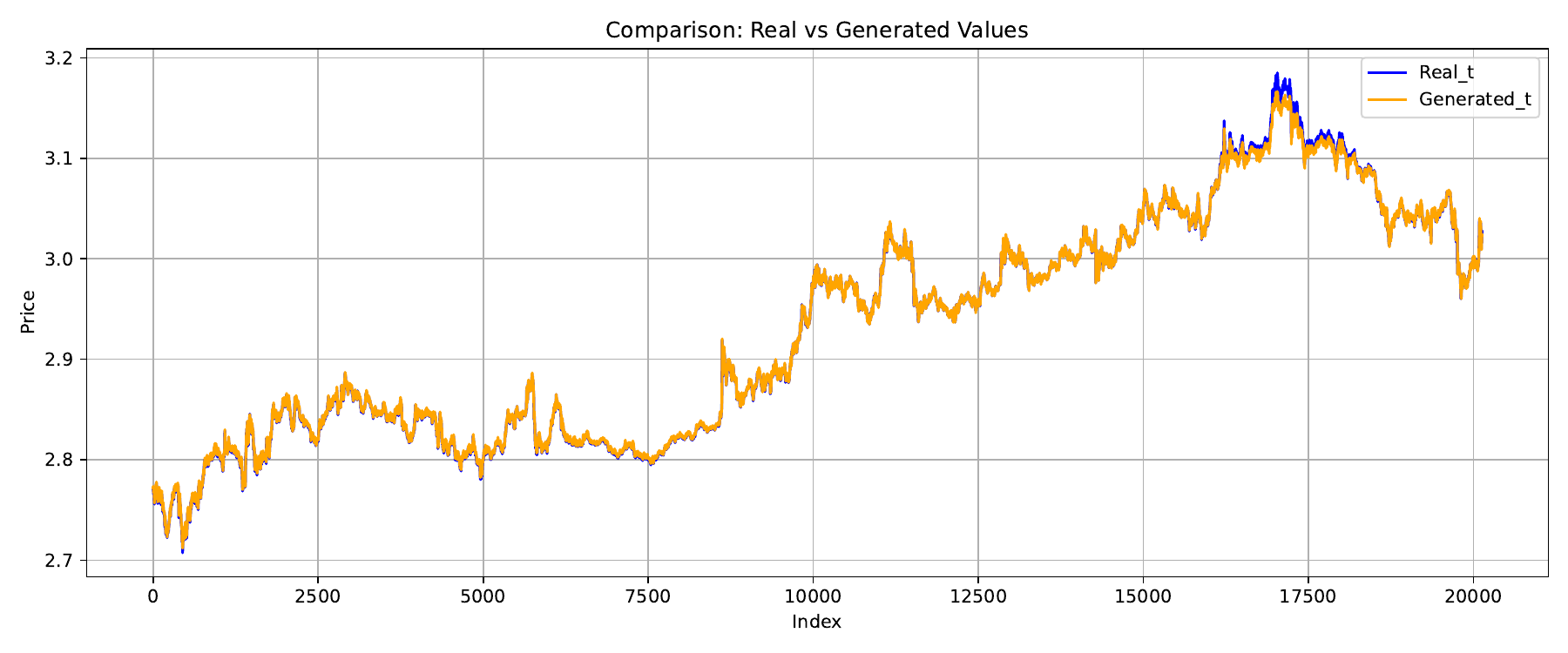}
\caption{XRP - data third period.}
\label{fig:XRPdata3}
\end{subfigure}

\begin{subfigure}{0.65\textwidth}
\includegraphics[width=\textwidth]{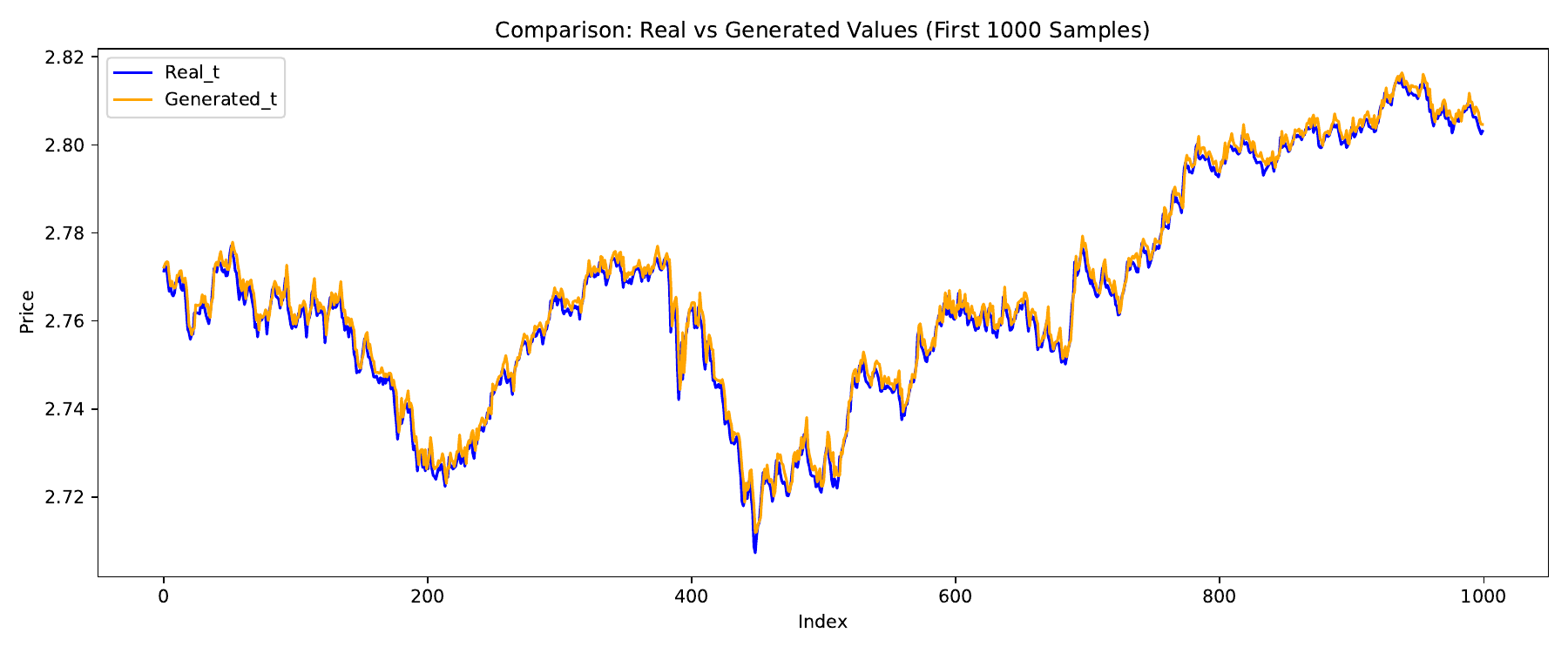}
\caption{XRP - data third period - first 1000 samples.}
\label{fig:XRPdata3-1000samples}
\end{subfigure}

\caption{Original series vs generated series: XRP - data third period.}
\label{fig:XRPdata3-result}
\end{figure}

The Fig.~\ref{fig:XRPdata3-result} shows a comparison between the generated and actual values for the third period of XRP values, and a sample of the first 1000 samples for better visualization.

\vspace*{-0.5cm}

The graphical visualization of the generated data behavior indicates that the developed model captured the closing pattern of BTC, ETH, and XRP during the testing phase. In the architecture, a range of 60 previous observations was defined, and normalization was performed using StandardScaler. The synthetic sequences faithfully reproduce the overall price slope, market oscillations, and point fluctuations of the original data. This shows that the network assimilated the fundamental conditional dependencies between the closing prices of the analyzed cryptocurrencies.

Regarding the learning phase, the configuration of the Adam optimizer (with parameters $\beta_1 = 0.5$ and $\beta_2 = 0.999$), the learning rate fixed at $2 \times 10^{-4}$, and the use of the BCEWithLogitsLoss function were crucial to maintaining the balance of the adversarial training, preventing critical failures in the generator. Finally, mini-batch processing ensured the necessary computational efficiency and statistical accuracy in creating the validated synthetic samples.

\subsection{Discussion}

Although the three cryptocurrencies analyzed exhibit similar general dynamics, the results show that the model's effectiveness varies consistently between BTC, ETH, and XRP, reflecting structural and statistical differences inherent to each time series.

In the case of BTC, synthetic data exhibits high agreement with the trajectory observed in real data, preserving both medium- and long-term trends as well as transitions between market regimes. This superior performance can be attributed to the greater liquidity, maturity, and informational efficiency of the BTC market, characteristics that contribute to time series with less relative noise and more stable patterns. Empirical studies indicate that Bitcoin exhibits greater statistical predictability when compared to other cryptocurrencies, which favors the learning of temporal dependencies by recurrent architectures, such as LSTMs, especially when integrated with adversarial models \cite{Urquhart2016, Katsiampa2017, LahmiriBekiros2019}.

Regarding ETH, the model is able to adequately capture the general direction of price movement; however, a systematic attenuation is noted in volatility peaks, with the generated series showing amplitudes lower than those observed in the real data. This behavior is consistent with results reported in the literature on GANs applied to financial series, in which the generator tends to smooth extreme events due to the difficulty in reproducing abrupt shocks and distributions with heavy tails \cite{Wiese2020QuantGAN, Yoon2019TimeGAN}. Furthermore, ETH is strongly influenced by exogenous factors, such as protocol updates, changes in the consensus mechanism, and the evolution of its smart contract ecosystem, which introduces additional non-stationarity and increases the complexity of the temporal modeling process.

In the case of XRP, the results indicate good local accuracy in certain intervals, although accompanied by greater sensitivity to short-term noise. This greater variability may be associated with the more speculative nature of the asset and its dependence on external events, such as regulatory announcements, judicial decisions, and institutional actions, which are not fully reflected in the price history. Previous studies highlight that crypto assets with lower market efficiency and high statistical nonlinearity pose additional challenges to the generalization capacity of neural network-based models, including GANs, making them more susceptible to spurious fluctuations \cite{LahmiriBekiros2019, Zhang2021CryptoDL}.

The choice of the StandardScaler method proves particularly suitable in the context of training GANs applied to financial time series. In gradient-based architectures, such as GANs and LSTM-type recurrent networks, data standardization directly contributes to the numerical stability of the optimization process, reducing problems associated with gradient explosion or disappearance.

Furthermore, standardization by mean and variance preserves the relative relationships between the values of the time series, an essential aspect for modeling dynamic dependencies over time. Unlike interval scaling methods, such as Min-Max Scaling, StandardScaler is less sensitive to extreme values, a relevant characteristic in financial series, which frequently exhibit heavy tails and highly volatile events.

Previous studies demonstrate that the use of standardized data favors the convergence of adversarial training and improves the generator's ability to capture complex statistical patterns, especially when combined with recurrent and conditional architectures \cite{goodfellow2014, Yoon2019TimeGAN, Wiese2020QuantGAN}. Therefore, the use of StandardScaler contributes to more stable and efficient learning, being a methodological choice consistent with the specialized literature.

In the context of GANs, the discriminator is modeled as a binary classifier responsible for distinguishing real samples from those synthesized by the generator. The objective function originally proposed by \cite{goodfellow2014} is explicitly based on maximizing binary cross-entropy, which makes BCE a conceptually aligned choice with the adversarial problem.

The use of BCEWithLogitsLoss, in particular, presents significant advantages over the separate application of the sigmoid function followed by binary cross-entropy. Firstly, the integration of these operations improves numerical stability during training, reducing the occurrence of zero or explosive gradients — a recurring problem in GANs, especially when the discriminator becomes overconfident. This feature ensures that informative gradients continue to be propagated to the generator through the backpropagation process, preserving the dynamics of the minimax game \cite{goodfellow2014, radford2016dcgan}.

Furthermore, by operating directly on the logits, BCEWithLogitsLoss allows the discriminator to maintain a richer representation of the uncertainty associated with its predictions, which contributes to more balanced training. Empirical studies, such as the work of \cite{radford2016dcgan} in the context of DCGAN, demonstrate that this formulation favors stable convergence and reduces the probability of mode collapse, especially when combined with appropriate adjustments of the optimizer and the learning rate.

In applications involving financial time series, characterized by high volatility and non-stationarity, the numerical robustness of BCEWithLogitsLoss becomes even more relevant. The function allows the discriminator to respond continuously to small statistical variations in the data, providing more consistent error signals to the generator during adversarial training, as discussed in modern GAN approaches for time series \cite{esteban2017realvalued, Yoon2019TimeGAN}.

\section{Conclusion} \label{sec:conclusion}

Our analysis demonstrates that a CGAN, equipped with an LSTM generator and a time window of 60 observations, effectively replicates the price dynamics of various cryptocurrencies, namely is capable of producing synthetic series. The model performed best on mature, liquid assets like Bitcoin (BTC), where temporal patterns are relatively stable. Conversely, performance dipped when modeling more volatile assets such as Ethereum and XRP. These assets' high structural volatility and susceptibility to exogenous factors highlight the ongoing challenge of modeling non-stationary financial series.

The results obtained reinforce the relevance of fundamental methodological choices, such as the appropriate definition of the time window, the adoption of appropriate normalization techniques, and the careful balancing of adversarial training, to ensure stability and convergence in GAN applications to financial time series. Additionally, the analysis highlights that behavioral and statistical differences between assets directly influence the model's ability to generalize, indicating that adaptive or asset-specific approaches may represent a promising path for future work.

\section*{Acknowledgments}
This work has been partially supported by QuIIN - EMBRAPII CIMATEC Competence Center in Quantum Technologies, with financial resources from the PPI IoT/Manufatura 4.0 / PPI HardwareBR of the MCTI grant number 053/2023, signed with EMBRAPII. The authors would like to thank the Supercomputing Center for Industrial Innovation (CS2I), the Reference Center for Artificial Intelligence (CRIA), and the Latin American Quantum Computing Center (LAQCC), all from SENAI CIMATEC, for providing all the technical and infrastructure support. We also thank the National Council for Scientific and Technological Development (CNPq, Brazil) for partially funding this work. Oscar M. Granados received funds for Universidad Jorge Tadeo Lozano (Grant 13456 for the Center on Intelligence and Innovation).

\bibliographystyle{unsrt}  
\bibliography{references}  


\end{document}